%% file: main.tex
\pdfoutput=1

\documentclass[11pt]{article}

\usepackage{acl}

\usepackage{times}
\usepackage{latexsym}
\usepackage{booktabs} 

\usepackage[T1]{fontenc}

\usepackage[utf8]{inputenc}

\usepackage{microtype}

\usepackage{inconsolata}
\usepackage{graphicx} 
\usepackage{algpseudocode}
\usepackage{algorithm}
\usepackage{caption}
\usepackage{subcaption}
\usepackage[export]{adjustbox} 
\usepackage{multirow} 
\usepackage{cleveref}
\input{macros}

\newcommand*\samethanks[1][\value{footnote}]{\footnotemark[#1]}

\title{What explains the success of cross-modal fine-tuning with ORCA?} 

\author{Paloma García-de-Herreros\thanks{~~Equal contribution.}\textsuperscript{1} \,
 Vagrant Gautam\samethanks\textsuperscript{1} \, Philipp Slusallek\textsuperscript{1, 2} \\ \textbf{Dietrich Klakow\textsuperscript{1}} \,
 \textbf{Marius Mosbach\textsuperscript{3,4}} \vspace{3px} \\ \vspace{3px}
 \textsuperscript{1}Saarland University~~~ 
\textsuperscript{2}DFKI~~~
 \textsuperscript{3}McGill University~~~
 \textsuperscript{4}Mila -- Quebec AI Institute\\
\small{\tt  \{pgherreros,vgautam\}@lsv.uni-saarland.de}
}

\begin{document}

\maketitle

\begin{abstract}
ORCA \citep{shen2023cross} is a recent technique for cross-modal fine-tuning,
i.e., applying pre-trained transformer models to modalities beyond their training data.
The technique consists primarily of training an embedder and fine-tuning the embedder and model.
Despite its high performance on a variety of downstream tasks, we do not understand precisely how each of these components contribute to ORCA's success.
Therefore, we run a series of ablations and find that embedder training does not help 2D tasks at all, contrary to what the original paper posits.
In 1D tasks, some amount of embedder training is necessary but more is not better.
In 4 out of 6 datasets we experiment with, it is model fine-tuning that makes the biggest difference.
Through our ablations and baselines, we contribute a better understanding of the individual components of ORCA.

\end{abstract}

\section{Introduction}

Modern AI is based on a pipeline of pre-training general-purpose models on vast amounts of data and then adapting them to specific tasks.
Examples across natural language processing (NLP) and computer vision (CV) typically focus on within-modality adaptation across, e.g., tasks or domains, but there is also a recent line of work that looks at leveraging pre-trained models \textit{across} modalities, e.g., Frozen Pretrained Transformers (FPT) \cite{lu2021pretrained}, ORCA \cite{shen2023cross}, OmniPred \cite{song2024omnipred}, Unified PDE Solver (UPS) \cite{shen2024ups}, inter alia.

ORCA is a recent example of a method for cross-modal fine-tuning~\citep{shen2023cross}.
It consists of a three-phase pipeline, shown in \Cref{fig:ORCA}.
First, a pre-trained transformer is selected, and a custom embedder and predictor are created to support any combination of input-output dimensions. Second, the embedder is trained to minimize the distance between a target and a proxy dataset, in order to map the target dataset into the embedding space of the pre-trained model. Finally, all three components are fine-tuned on data from the target task.

According to \citet{shen2023cross}, the reason for ORCA's success is the training of the custom embedder. We expand on their ablations to better understand the contributions of ORCA's individual components,
focusing on ablating the second and third stages of the pipeline. Our specific research questions are:

\begin{figure}[t]
    \centering
    \includegraphics[width=1\linewidth]{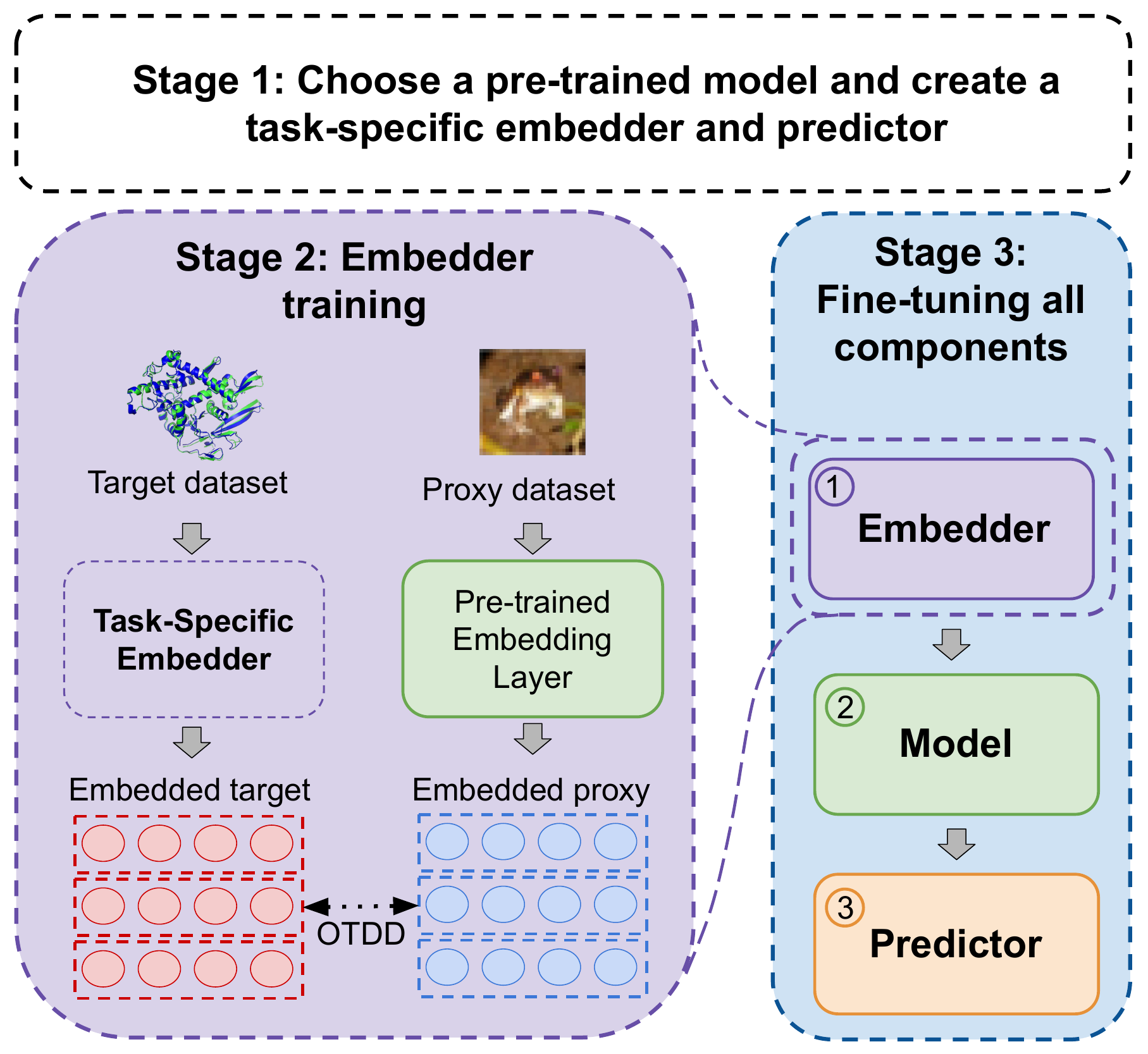}
    \caption{The ORCA pipeline. Stage 2 involves training the task-specific embedder. Stage 3 fine-tunes the embedder, the pre-trained encoder, and the predictor.}\vspace{-3mm}
    \label{fig:ORCA}
\end{figure}

\vspace{-1mm}
\begin{enumerate}
    \item How does the choice of proxy dataset affect performance? (\S\ref{sec:proxy-dataset-choice})
    \vspace{-2mm}
    \item Does doing (more) embedder training improve performance? (\S\ref{sec:otdd-performance})
    \vspace{-2mm}
    \item What do the embedder and the pre-trained model contribute individually? (\S\ref{sec:frozen})
    \vspace{-2mm}
    \item How much pre-training is necessary for cross-modal transfer? (\S\ref{sec:how-much-pretrained-knowledge})
\end{enumerate}
\vspace{-1mm}

By disentangling the contributions of embedder training and model fine-tuning, our results provide a more nuanced perspective on the success of cross-modal fine-tuning with ORCA. Additionally, our findings highlight the importance of strong baselines and careful ablations when making claims about \textit{why} a method works.

\begin{figure*}[ht]
     \centering
     \begin{subfigure}[b]{0.3\linewidth}
         \centering
         \includegraphics[width=\linewidth]{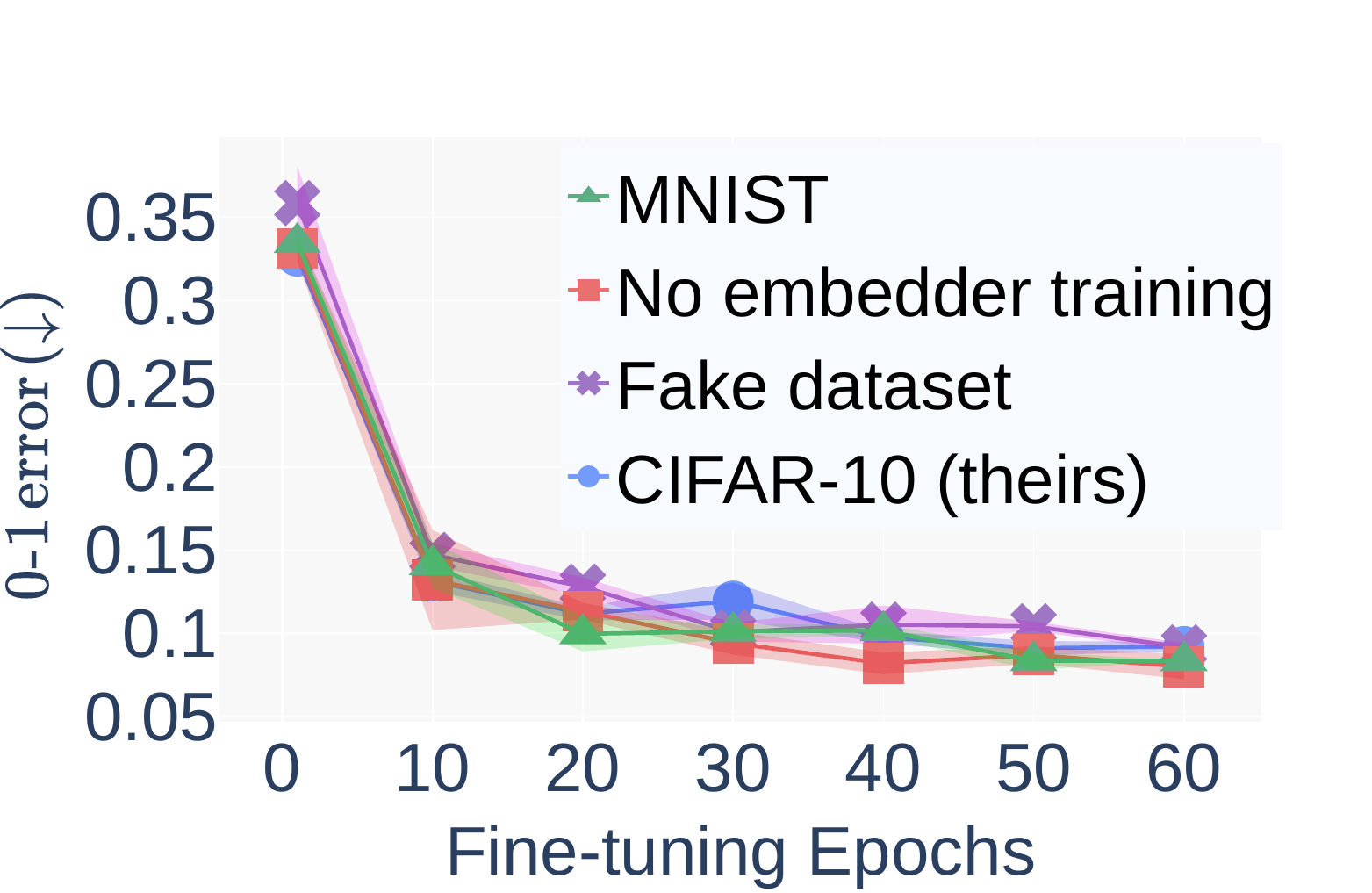}
         \caption{NinaPro}
         \label{fig:ninaproxy}
     \end{subfigure}
     \hfill
     \begin{subfigure}[b]{0.3\linewidth}
         \centering
         \includegraphics[width=\linewidth]{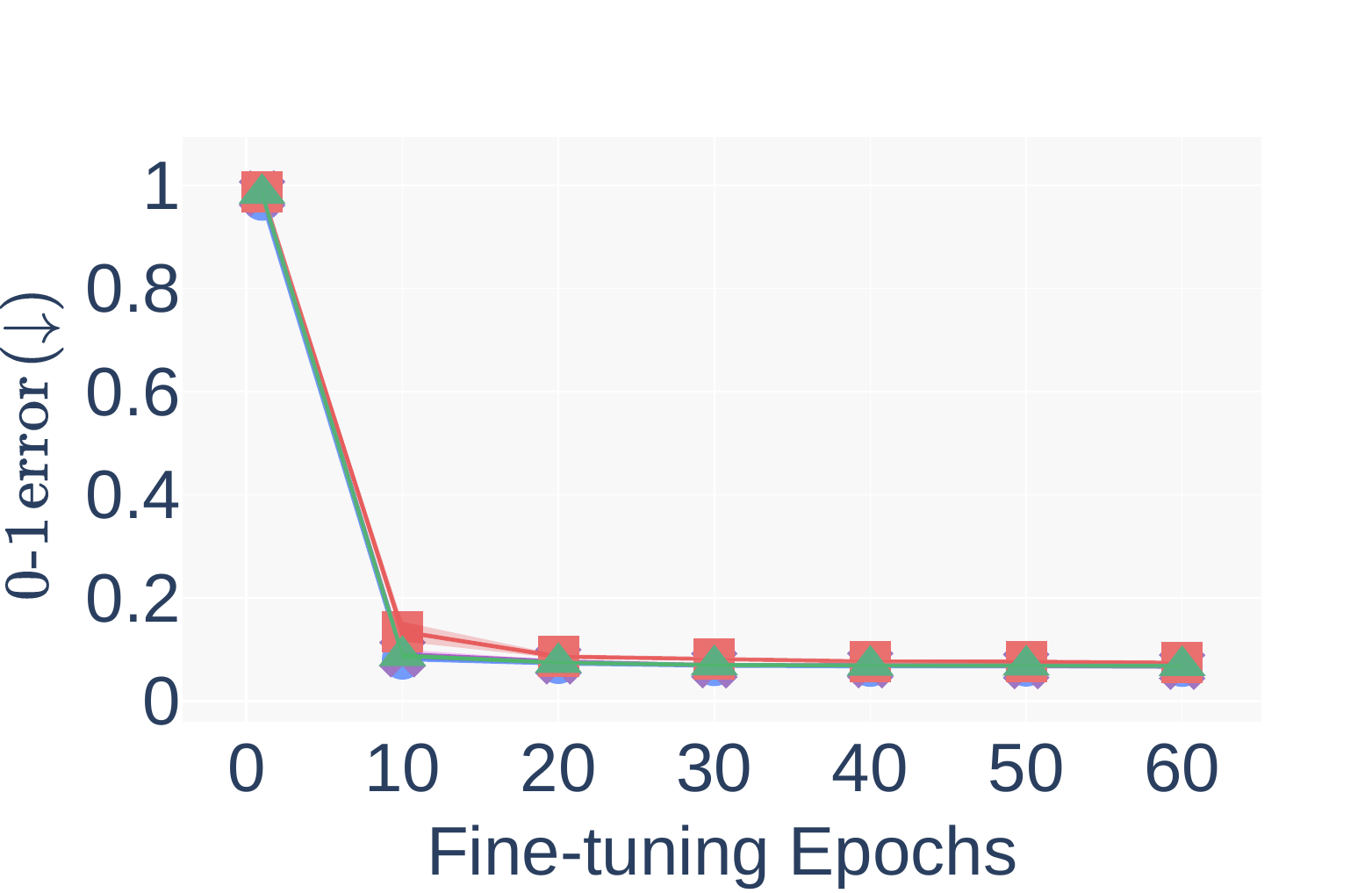}
         \caption{CIFAR-100}
         \label{fig:cifarproxy}
     \end{subfigure}
     \hfill
     \begin{subfigure}[b]{0.3\linewidth}
         \centering
         \includegraphics[width=\linewidth]{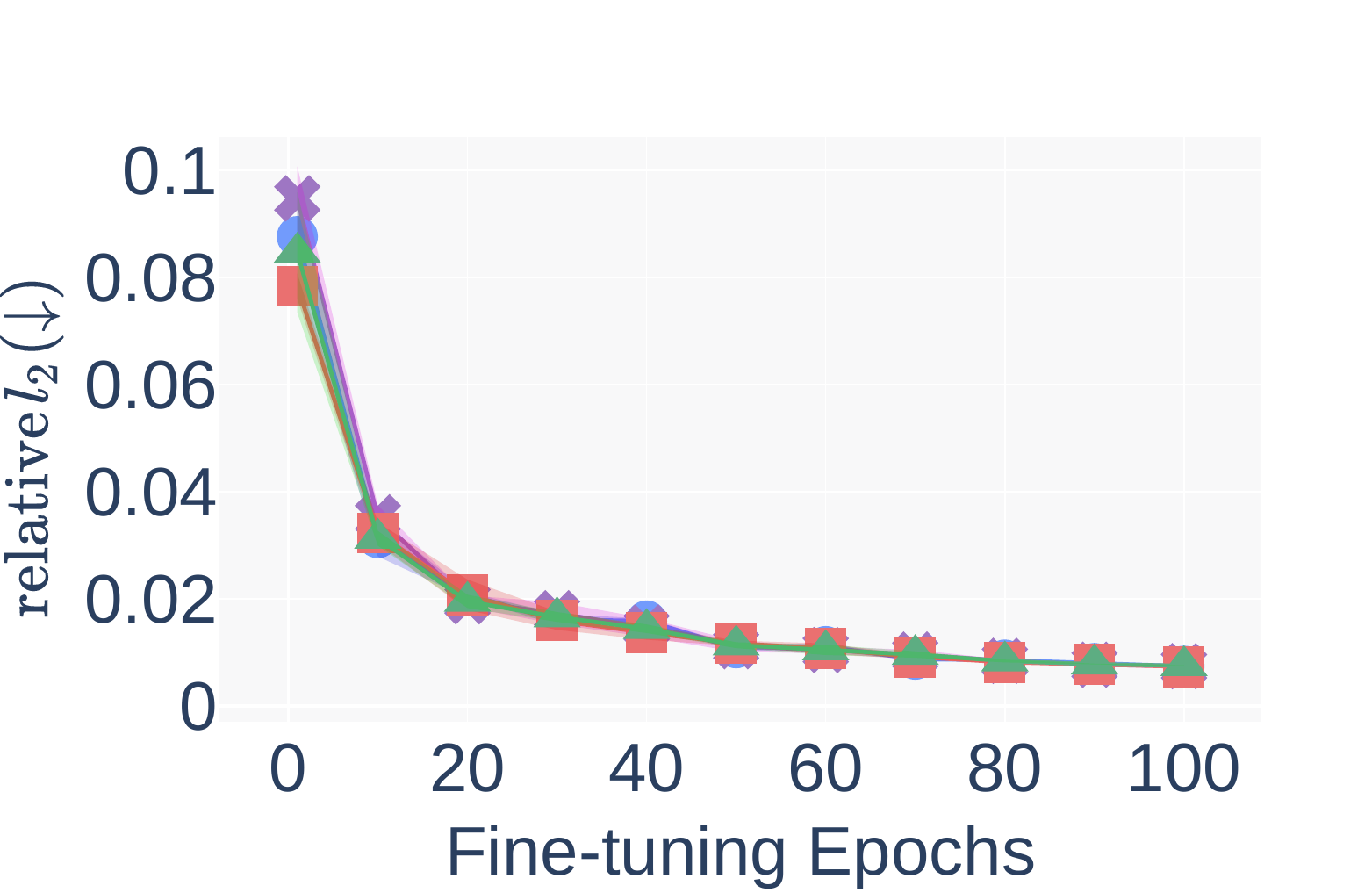}
         \caption{Darcy Flow}
         \label{fig:darcyproxy}
     \end{subfigure}
     \\
     \begin{subfigure}[b]{0.3\linewidth}
         \centering
         \includegraphics[width=\linewidth]{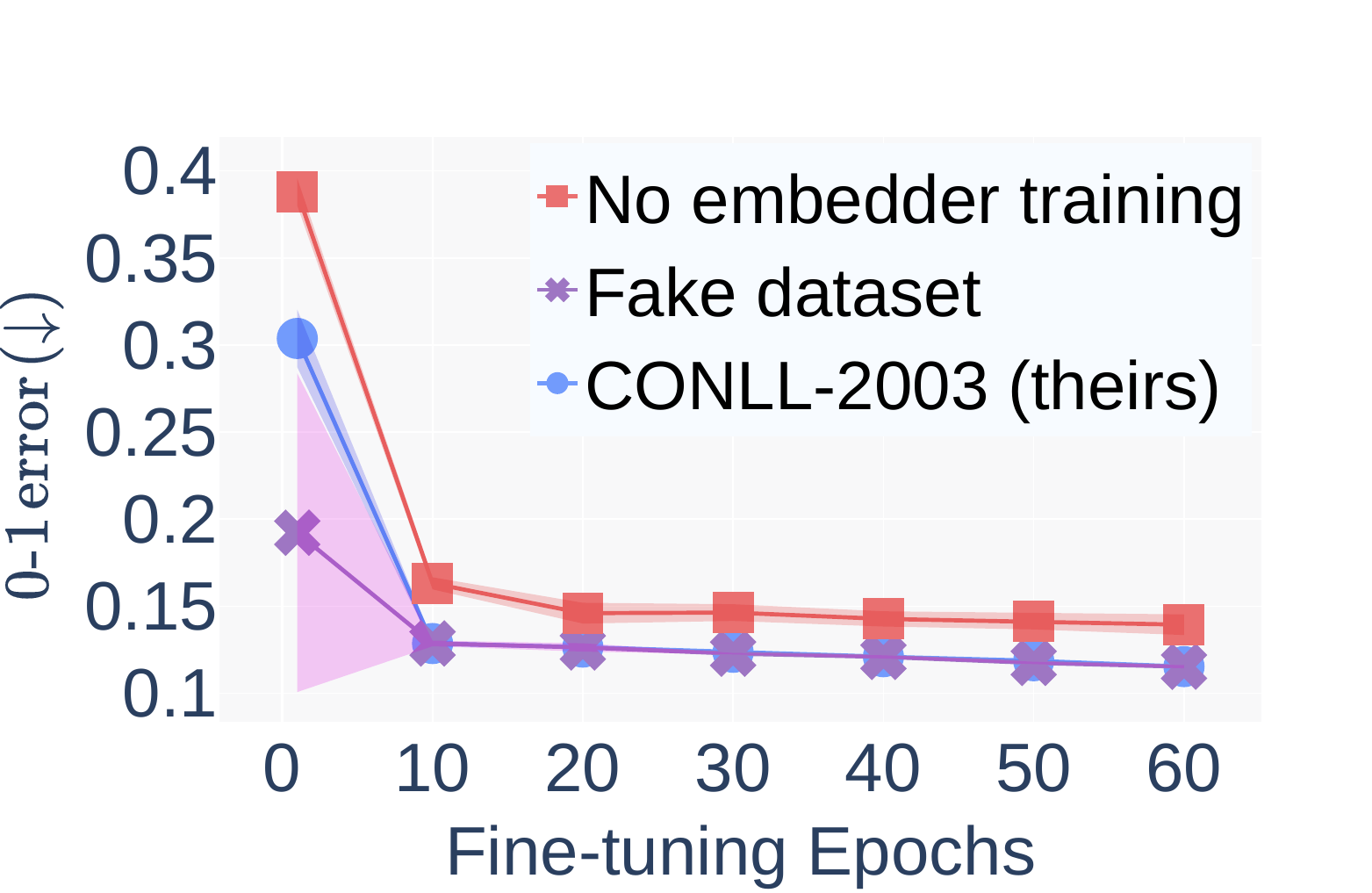}
         \caption{Satellite}
         \label{fig:Satelliteproxy}
     \end{subfigure}
     \hfill
     \begin{subfigure}[b]{0.3\linewidth}
         \centering
         \includegraphics[width=\linewidth]{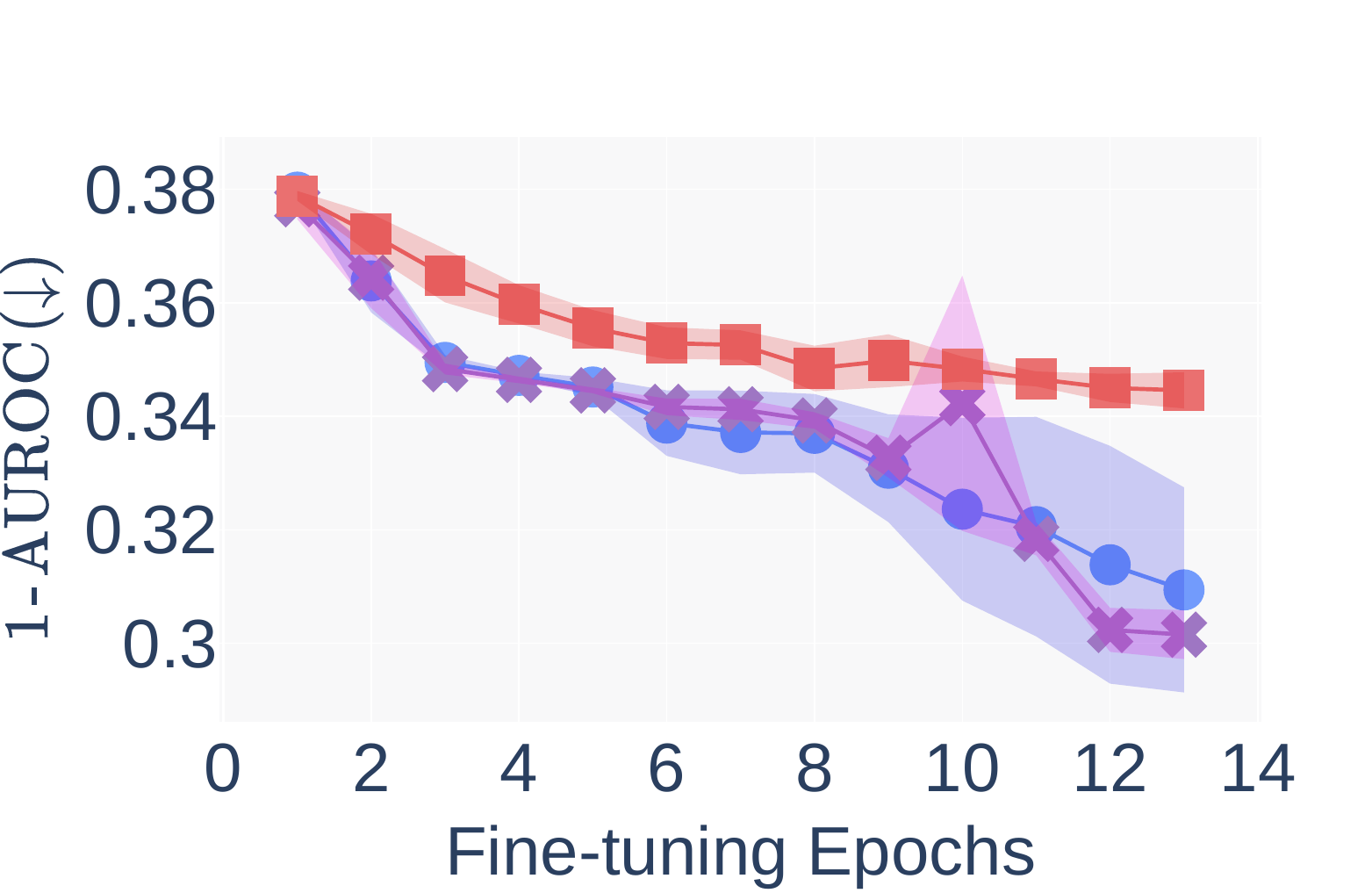}
         \caption{DeepSEA}
         \label{fig:deepseaproxy}
     \end{subfigure}
     \hfill
     \begin{subfigure}[b]{0.3\linewidth}
         \centering
         \includegraphics[width=\linewidth]{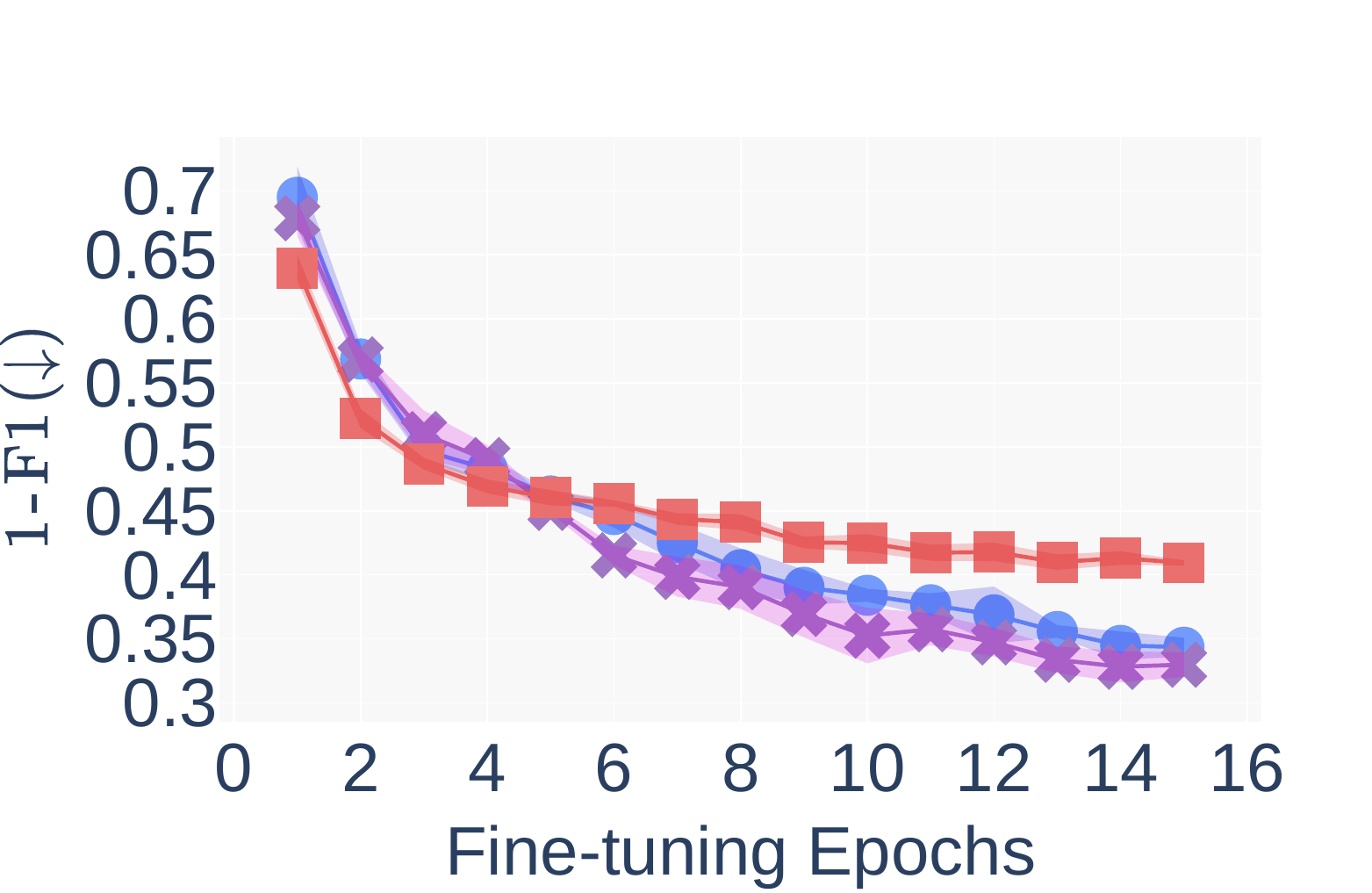}
         \caption{ECG}
         \label{fig:ecgproxy}
     \end{subfigure}
        \caption{Per-epoch fine-tuning performance on 2D tasks (above) and 1D tasks (below) when the embedder is trained with different proxy datasets or not trained at all, i.e., naive fine-tuning.}
        \label{fig:proxydatasets}
\end{figure*}

\section{Experimental setup}

Unless otherwise specified, we follow the ORCA paper in using RoBERTa-base \cite{DBLP:journals/corr/abs-1907-11692} and Swin-base \cite{liu2021swin} as the pre-trained transformers, a convolutional architecture for the embedder, and a linear transformation for the predictor (see \Cref{sec:appendix:embedder-details} for details).
We also use optimal transport dataset distance (OTDD; \citealp{alvarez2020geometric}) as the loss function during embedder training. All our experiments use their publicly available code.\footnote{\href{https://github.com/sjunhongshen/ORCA/}{https://github.com/sjunhongshen/ORCA/}} For training, we use the same hyperparameters as they do, except for the batch size when training on Satellite (64) and ECG (32) data. We evaluate on six target datasets that appear in the original paper, chosen to represent all pairs of dimensions and types, and we experiment with various proxy datasets. Dataset details are shown in \Cref{sec:dataset}.

\paragraph{Target datasets.} 

We select three 2D datasets (NinaPro, CIFAR-100, and Darcy Flow) and three 1D datasets (Satellite, DeepSEA, and ECG) from the NAS-Bench-360 benchmark \cite{tu2022bench}. 

\paragraph{Proxy datasets.}
The original paper uses CIFAR-10~\citep{Krizhevsky09learningmultiple} as the proxy dataset for all 2D tasks, and CoNLL 2003 \cite{tjong-kim-sang-de-meulder-2003-introduction} for all 1D tasks.
We experiment with additional proxy datasets to analyze their influence on overall performance.

For the 2D tasks, we compare to two other image datasets that maintain the same number of classes: MNIST \cite{deng2012mnist}, a different image dataset, and Fakedata\footnote{From torchvision.datasets.}, a dataset of randomly classified white noise images \cite{NEURIPS2019_9015}.

For the 1D tasks, we compare to a custom-created fake dataset classifying randomly generated language feature vectors into the same number of classes as CoNLL.

\begin{figure*}[h]
     \centering
        \begin{subfigure}[b]{0.3\linewidth}
         \centering
         \includegraphics[width=\linewidth]{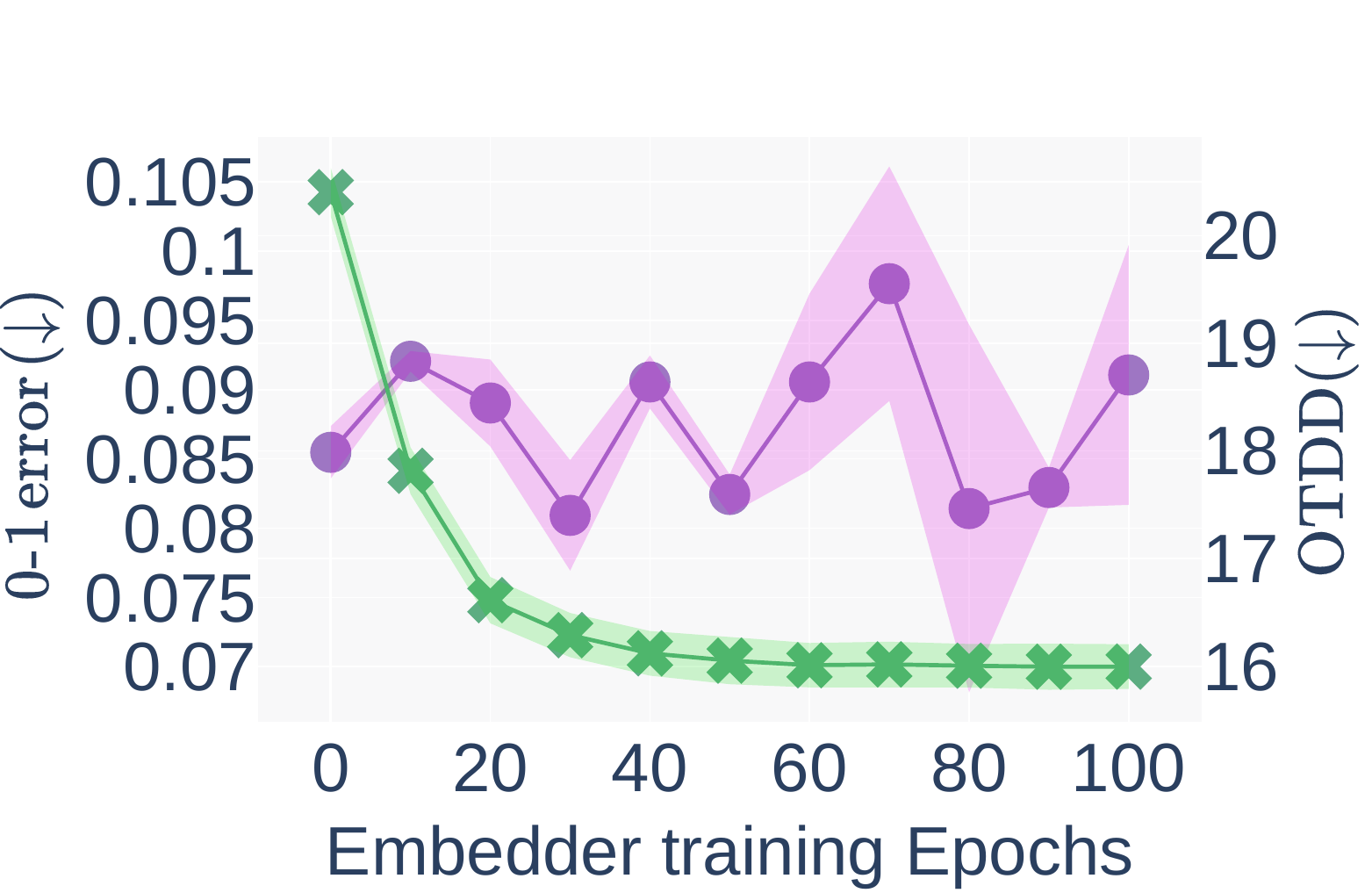}
         \caption{NinaPro}
         \label{fig:ninapro-otdd}
     \end{subfigure}
     \hfill
     \begin{subfigure}[b]{0.3\linewidth}
         \centering
         \includegraphics[width=\linewidth]{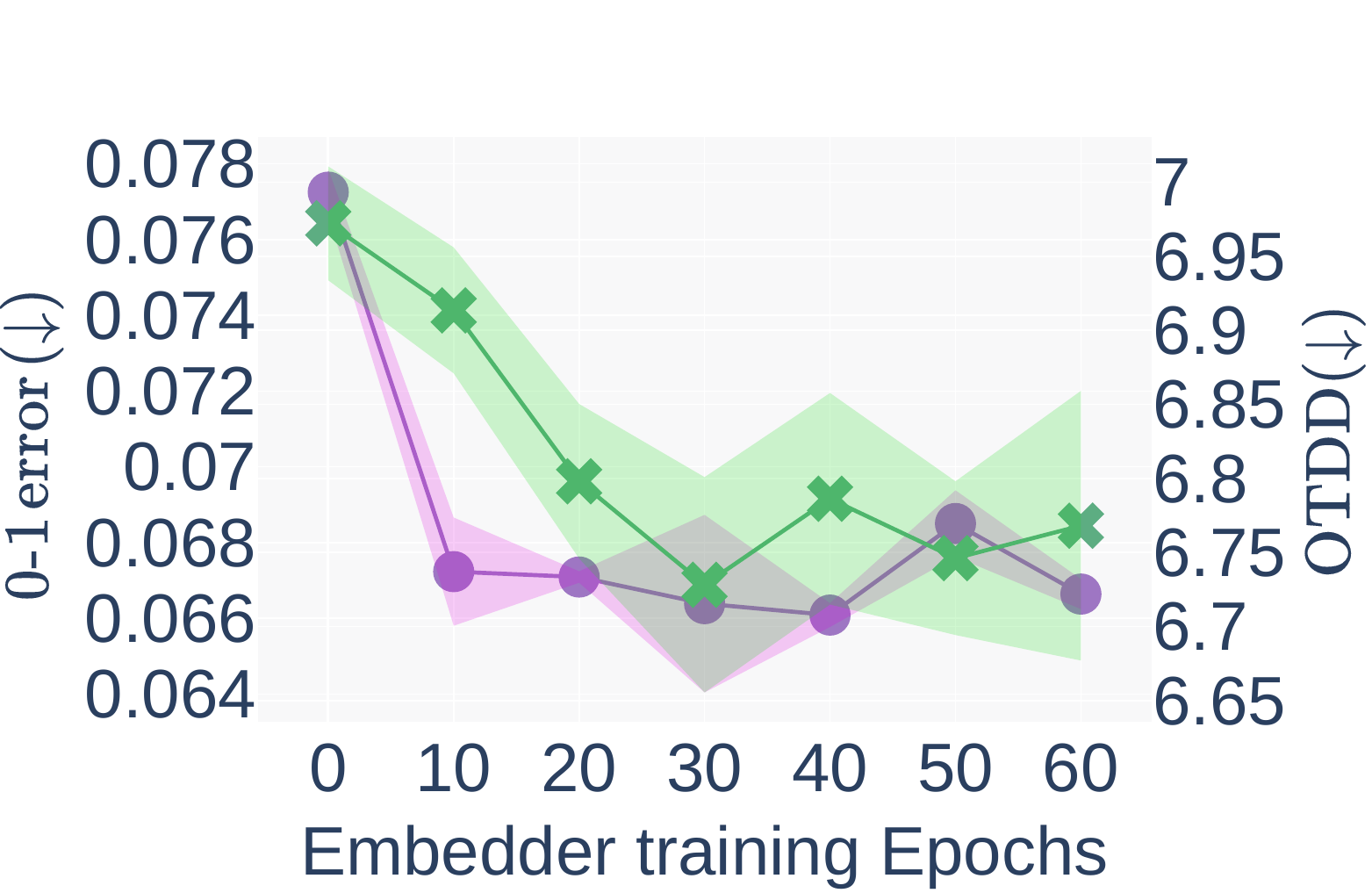}
         \caption{CIFAR-100}
         \label{fig:cifar-otdd}
     \end{subfigure}
     \hfill
     \begin{subfigure}[b]{0.3\linewidth}
         \centering
         \includegraphics[width=\linewidth]{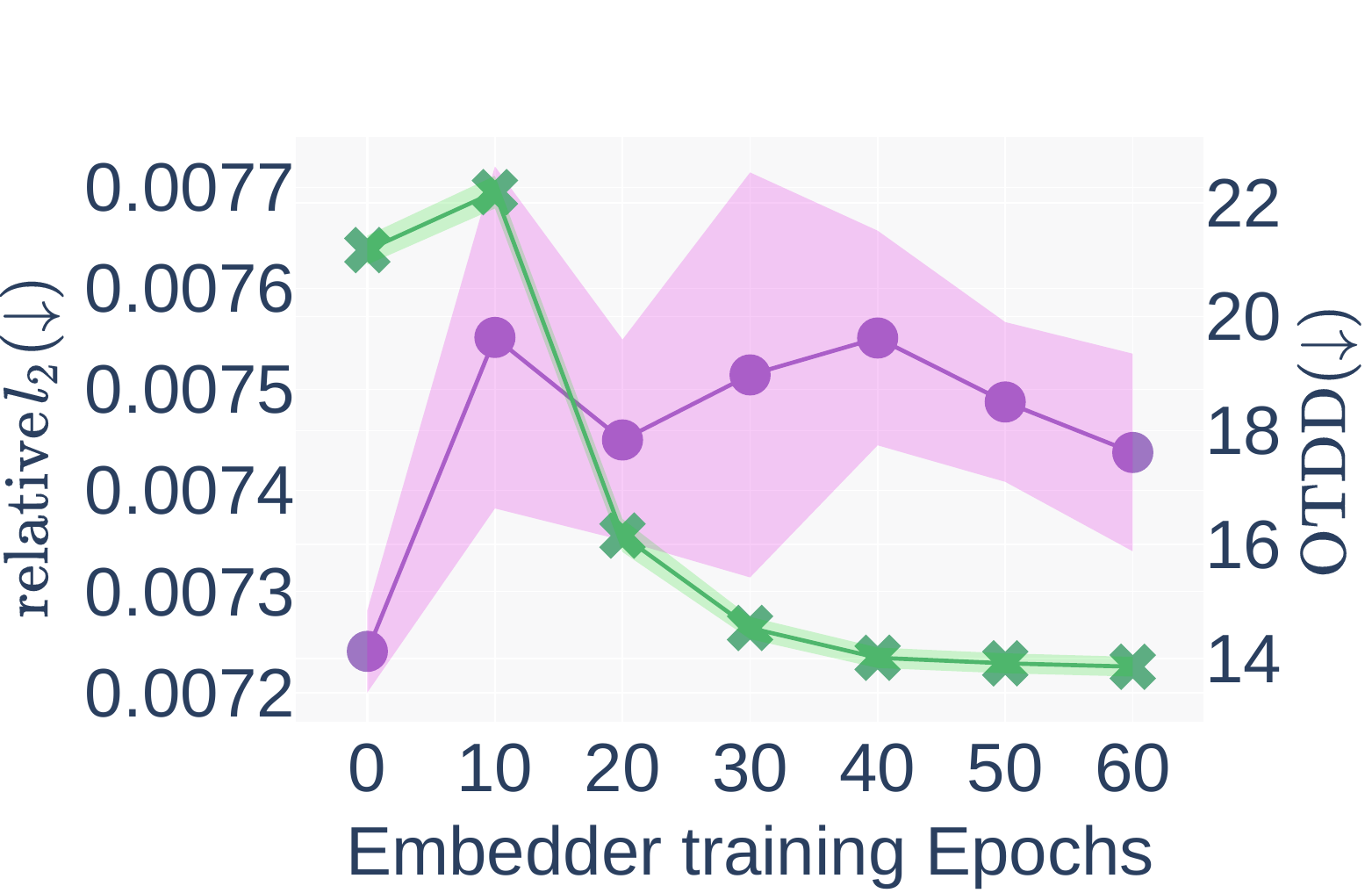}
         \caption{Darcy Flow}
         \label{fig:darcy-otdd}
     \end{subfigure}
          \begin{subfigure}[b]{0.3\linewidth}
         \centering
         \includegraphics[width=\linewidth]{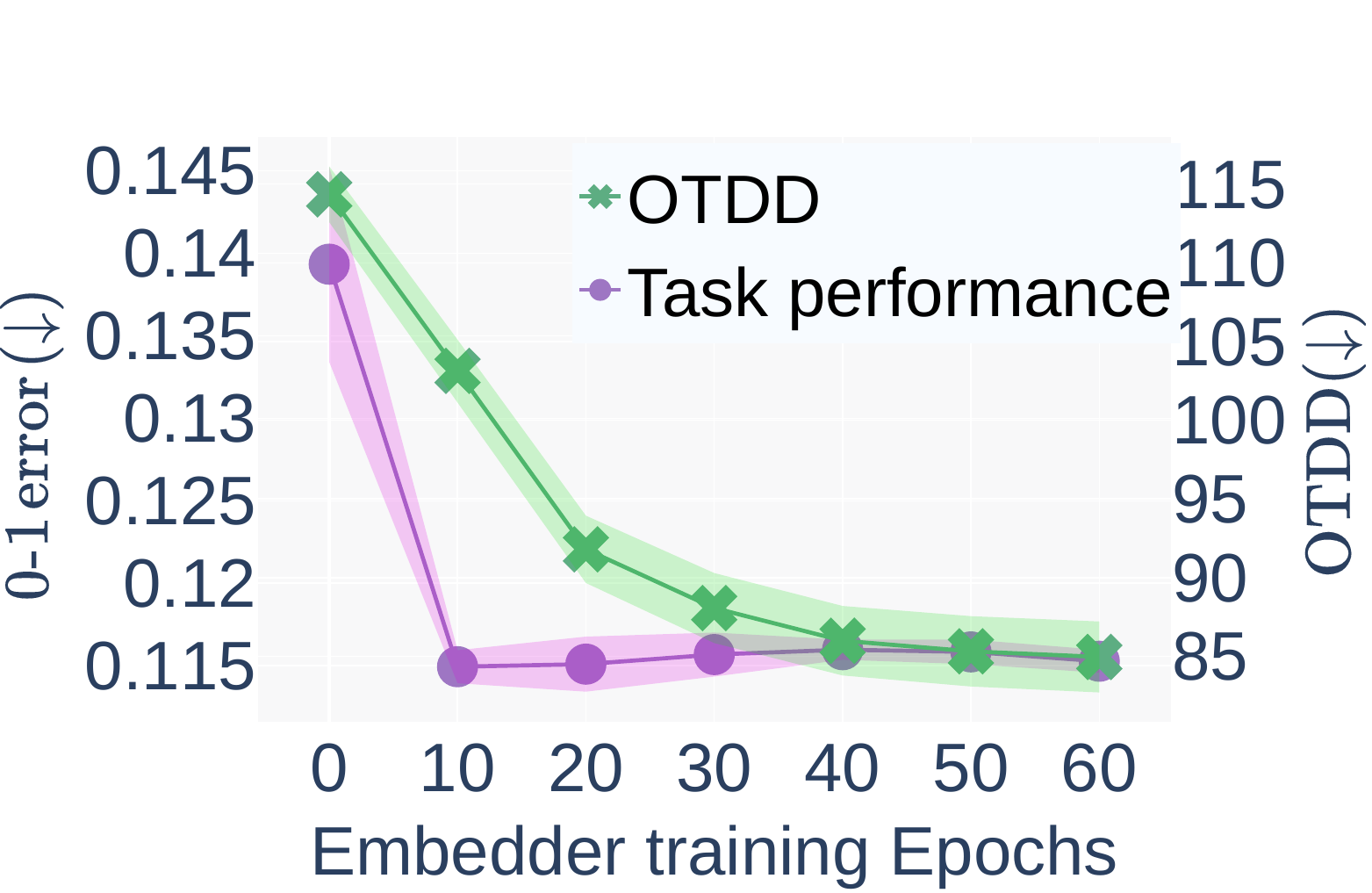}
         \caption{Satellite}
         \label{fig:satellite-otdd}
     \end{subfigure}
     \hfill
     \begin{subfigure}[b]{0.3\linewidth}
         \centering
         \includegraphics[width=\linewidth]{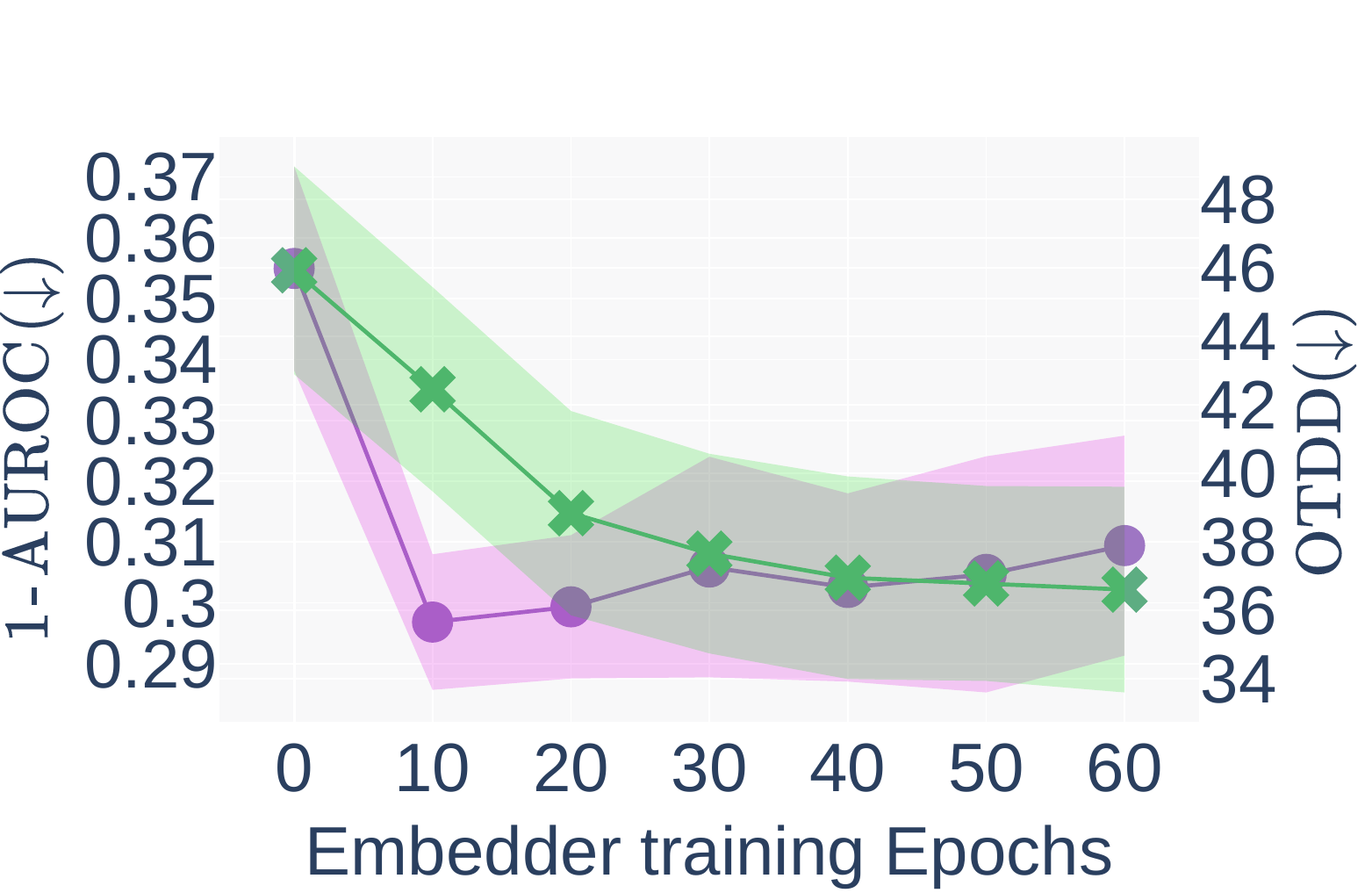}
         \caption{DeepSEA}
         \label{fig:deepsea-otdd}
     \end{subfigure}
          \hfill
     \begin{subfigure}[b]{0.3\linewidth}
         \centering
         \includegraphics[width=\linewidth]{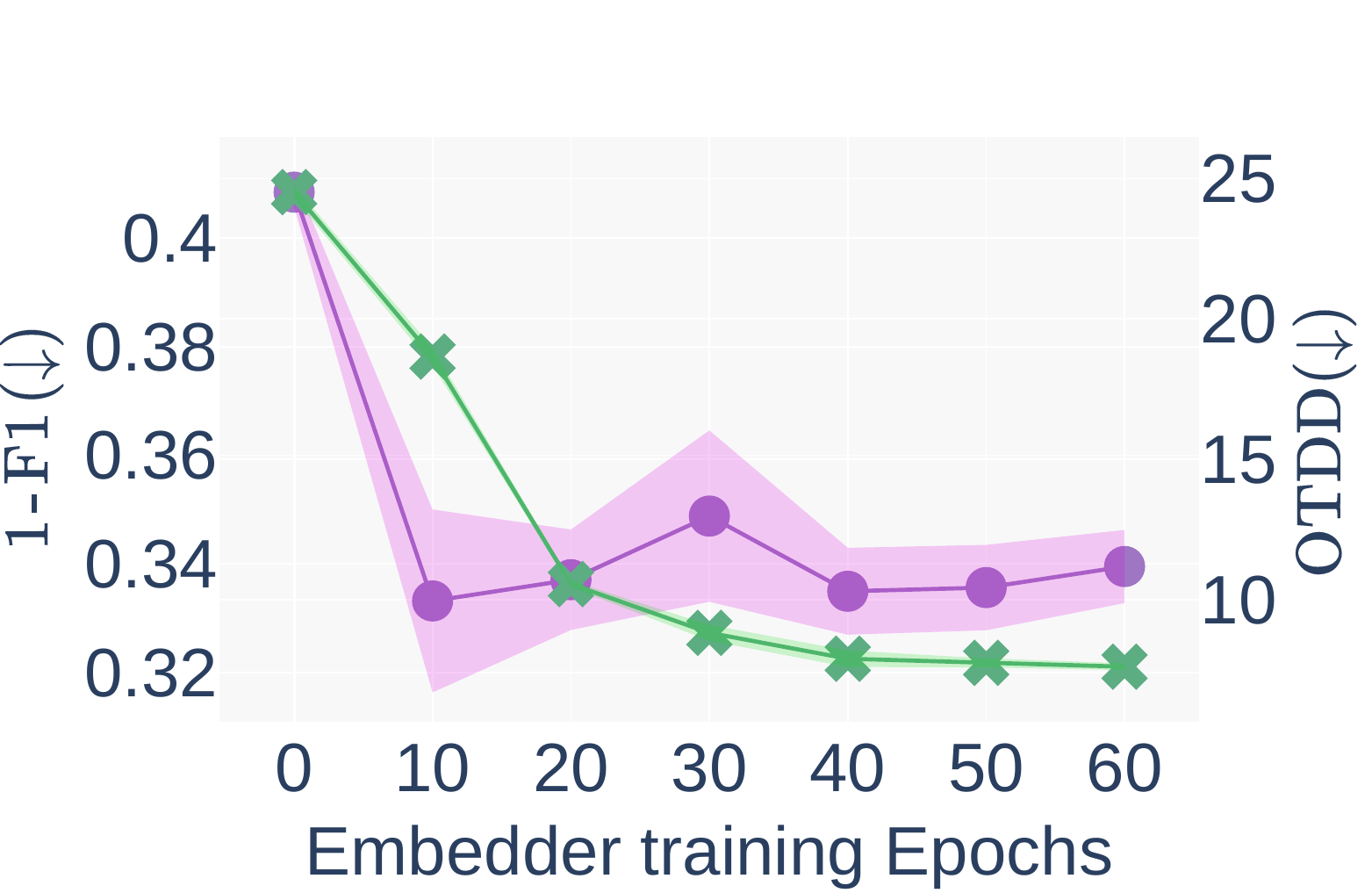}
         \caption{ECG}
         \label{fig:ecg-otdd}
     \end{subfigure}
        \caption{Per-epoch embedder training comparing OTDD ($\downarrow$) (metric minimized during this stage) to downstream task performance ($\downarrow$).}
        \label{fig:otdd}
\end{figure*}

\section{How does the choice of proxy dataset affect performance?}
\label{sec:proxy-dataset-choice}

In this section, we experiment with the choice of proxy dataset for the tasks. As a baseline, we compare to just fine-tuning the embedder, model and predictor, without training the embedder first.

As \Cref{fig:proxydatasets} shows, all fine-tuning curves for the 2D datasets (first row) overlap, indicating that the choice of proxy dataset is not important. Even fake data as a proxy dataset results in the same performance. Similarly, for the 1D tasks (second row), there is no real difference between using CoNLL and fake embeddings. Together, this shows that \textbf{the choice of proxy dataset for embedder training does not matter for ORCA to work}.

Comparing to a naive fine-tuning baseline allows us to evaluate the claim that ``ORCA consistently outperforms naive fine-tuning'' \citep{shen2023cross}. We find that \textbf{embedder training does play a role in the 1D tasks, but does not matter for 2D tasks}, even in the early stages of fine-tuning.

\section{(More) embedder training is not the secret to ORCA's success}
\label{sec:otdd-performance}

The previous results motivate us to more closely examine the role of embedder training in ORCA. In this stage, the OTDD metric is used to quantify the distance between the proxy and target embeddings. The authors minimize OTDD, claiming that ``as the dataset distance decreases, the fine-tuning accuracy increases'' \citep{shen2023cross}.

However, when we examine the relationship between OTDD and downstream task performance, we find that \textbf{embedder training is unnecessary in two out of six tasks} (Figures \ref{fig:ninapro-otdd} and \ref{fig:darcy-otdd}). For the remaining four tasks, \textbf{training the embedder more can even lead to worse task performance}.

As this section and the previous one show that embedder training does not affect final performance on the 2D tasks, we focus on the 1D tasks for our remaining experiments.

\section{Which components of ORCA are really necessary?}
\label{sec:frozen}

\begin{figure*}[ht] 
    \centering
  \begin{tabular}{lccc}
        & \hspace{4mm} \small{Satellite} & \hspace{4mm} \small{DeepSEA} & \hspace{4mm} \small{ECG}\vspace{1mm} \\
        & \multicolumn{3}{c}{\includegraphics[width=0.4\linewidth]{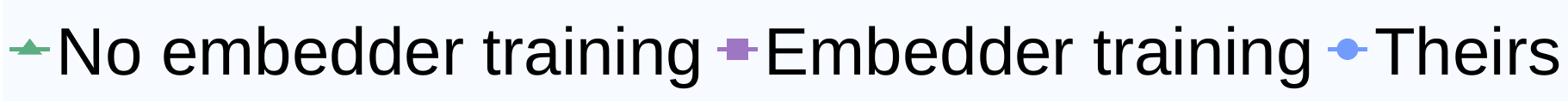}}\vspace{-2mm} \\
        \small{Freezing both} & \includegraphics[valign=m,width=0.23\linewidth]{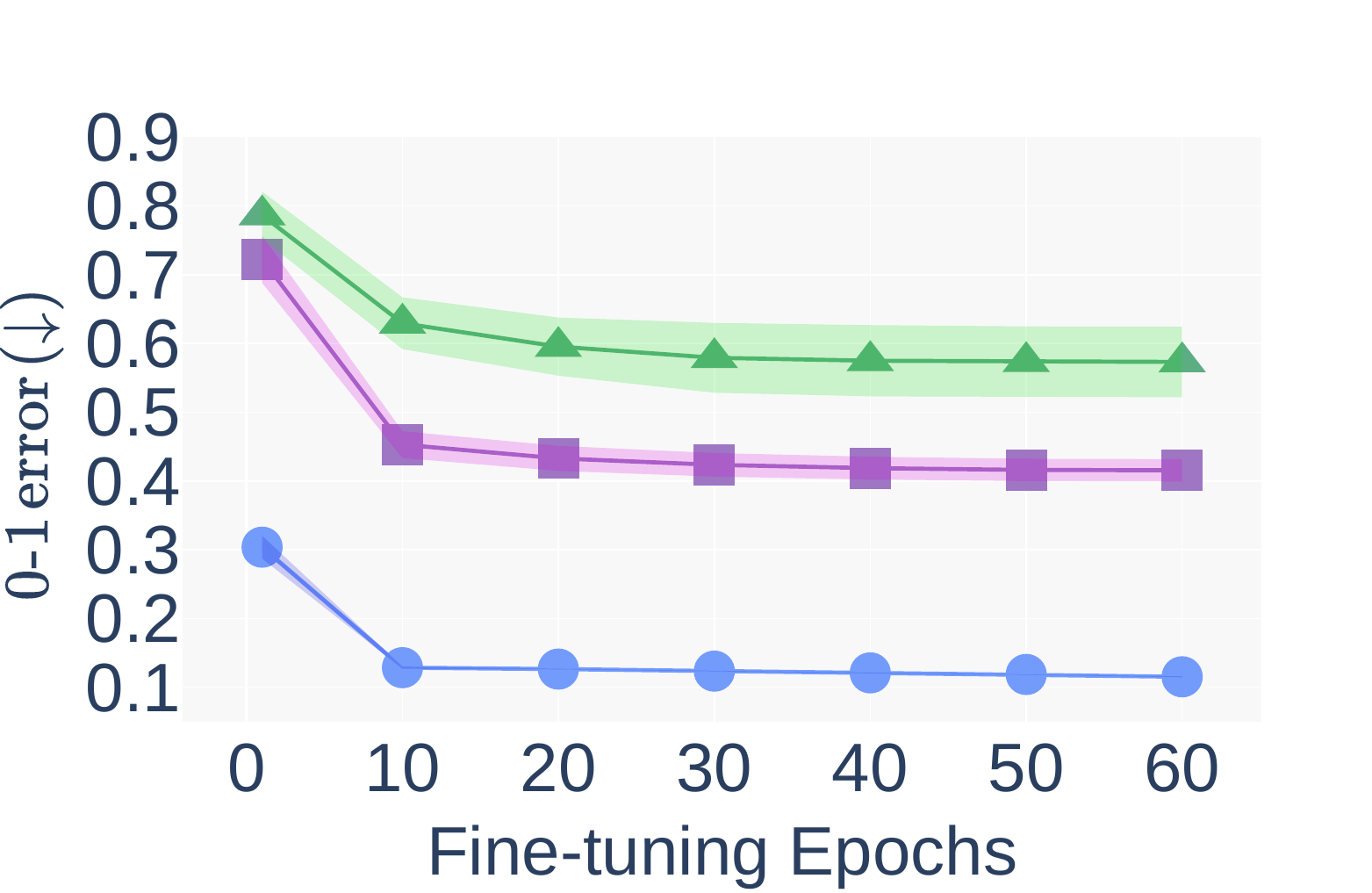} &
        \includegraphics[valign=m,width=0.23\linewidth]{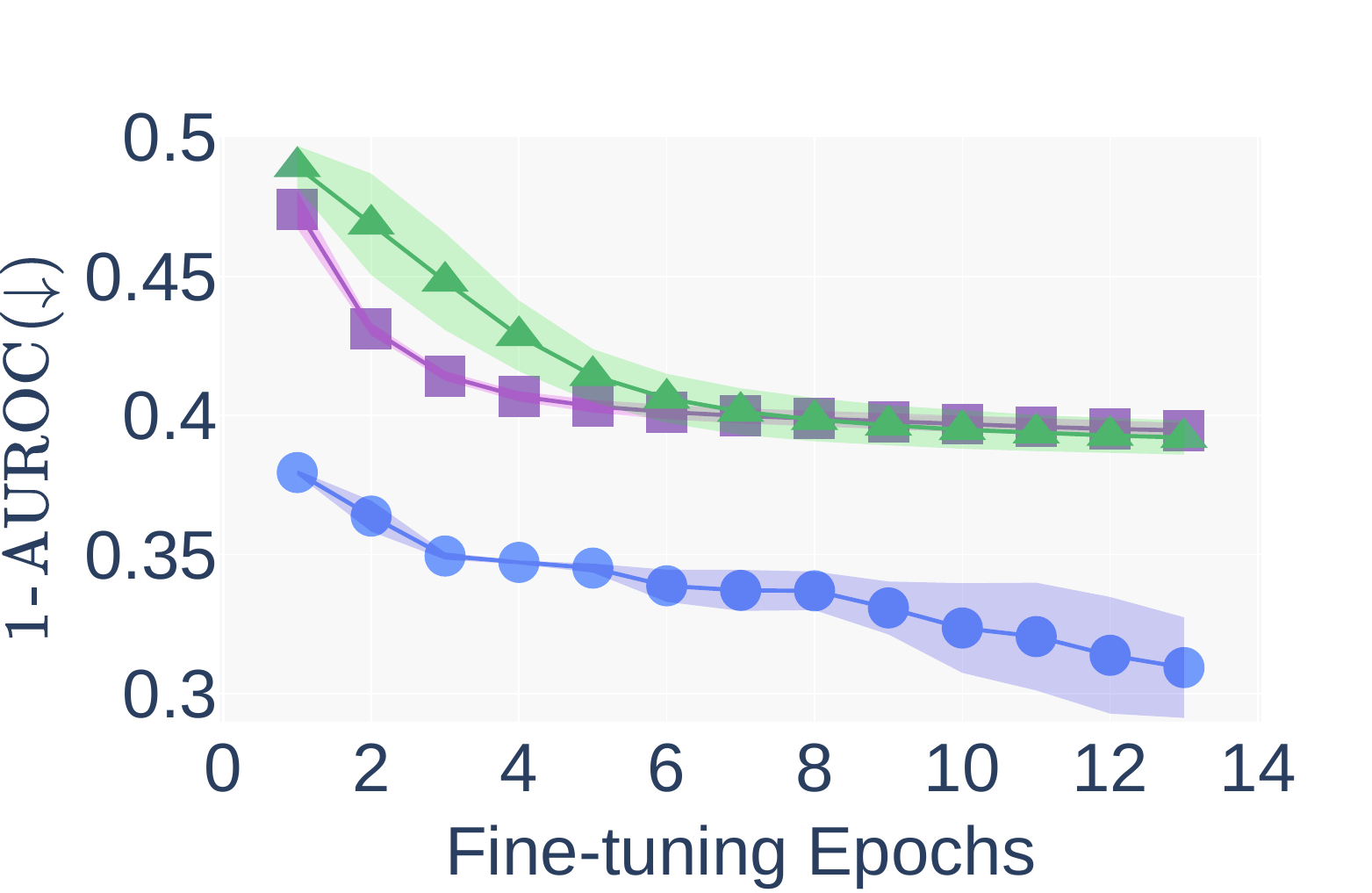} &
        \includegraphics[valign=m,width=0.23\linewidth]{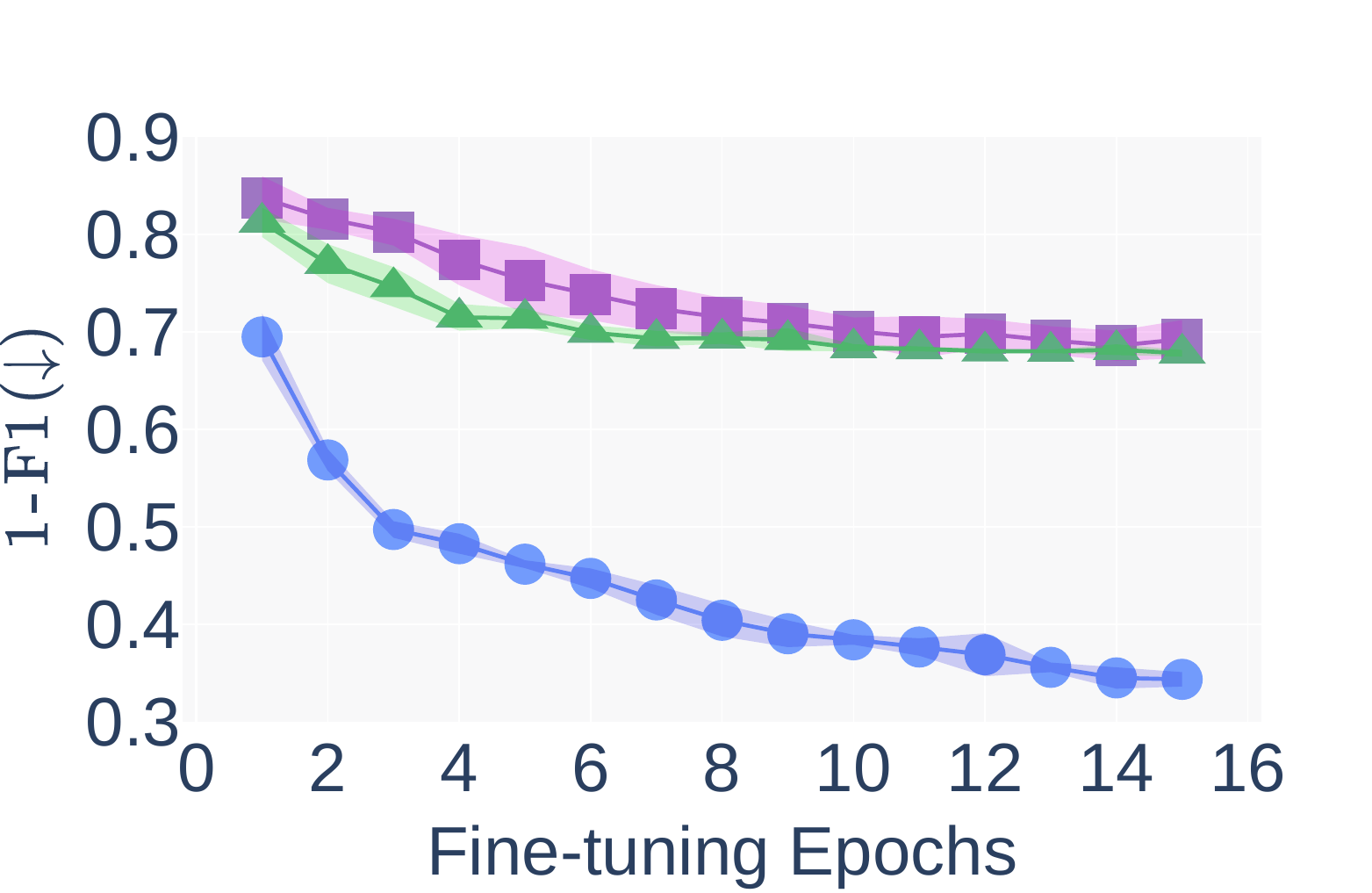} 
         \\
        \small{Freezing the model} & \includegraphics[valign=m,width=0.23\linewidth]{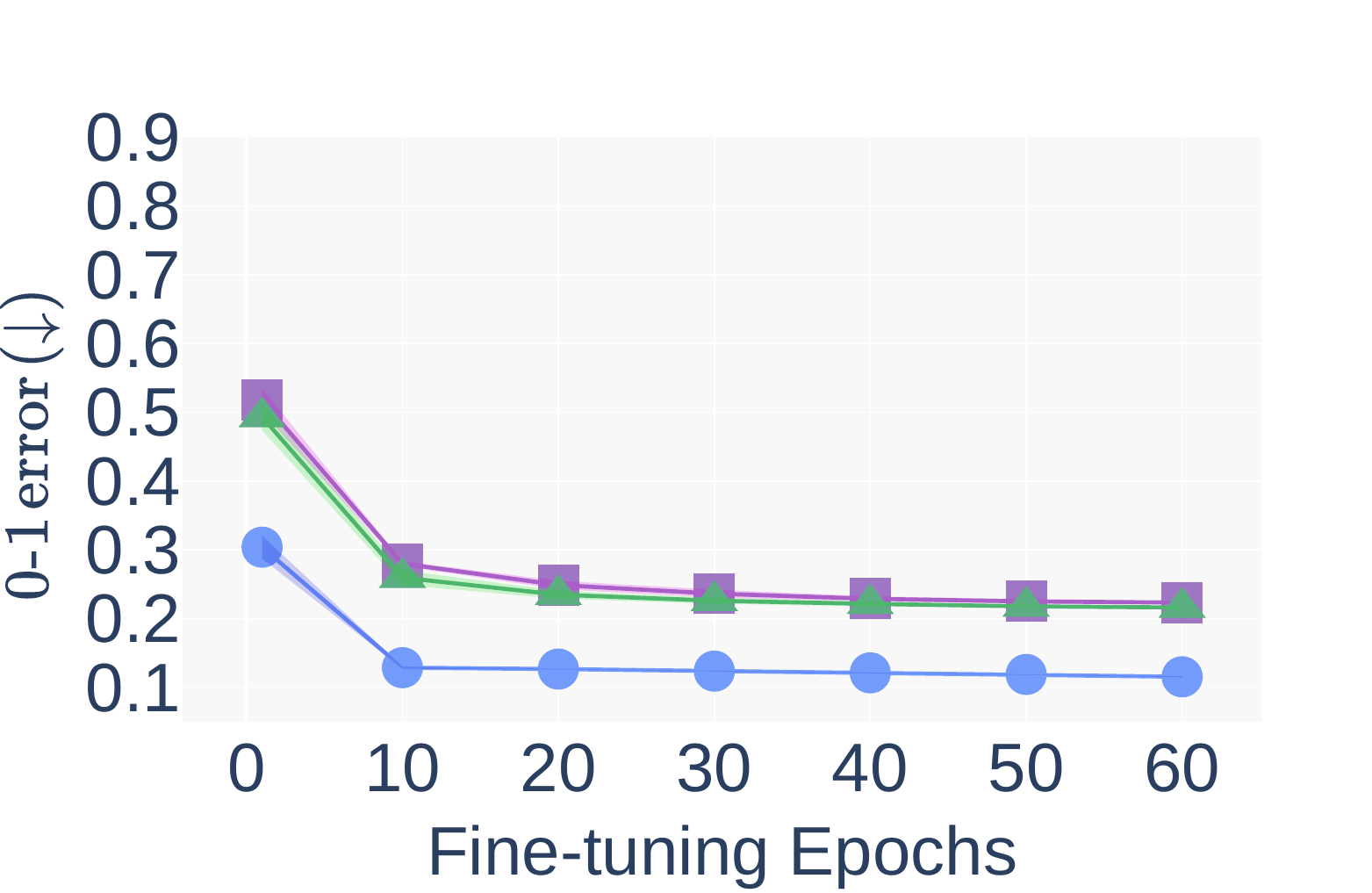} &
        \includegraphics[valign=m,width=0.23\linewidth]{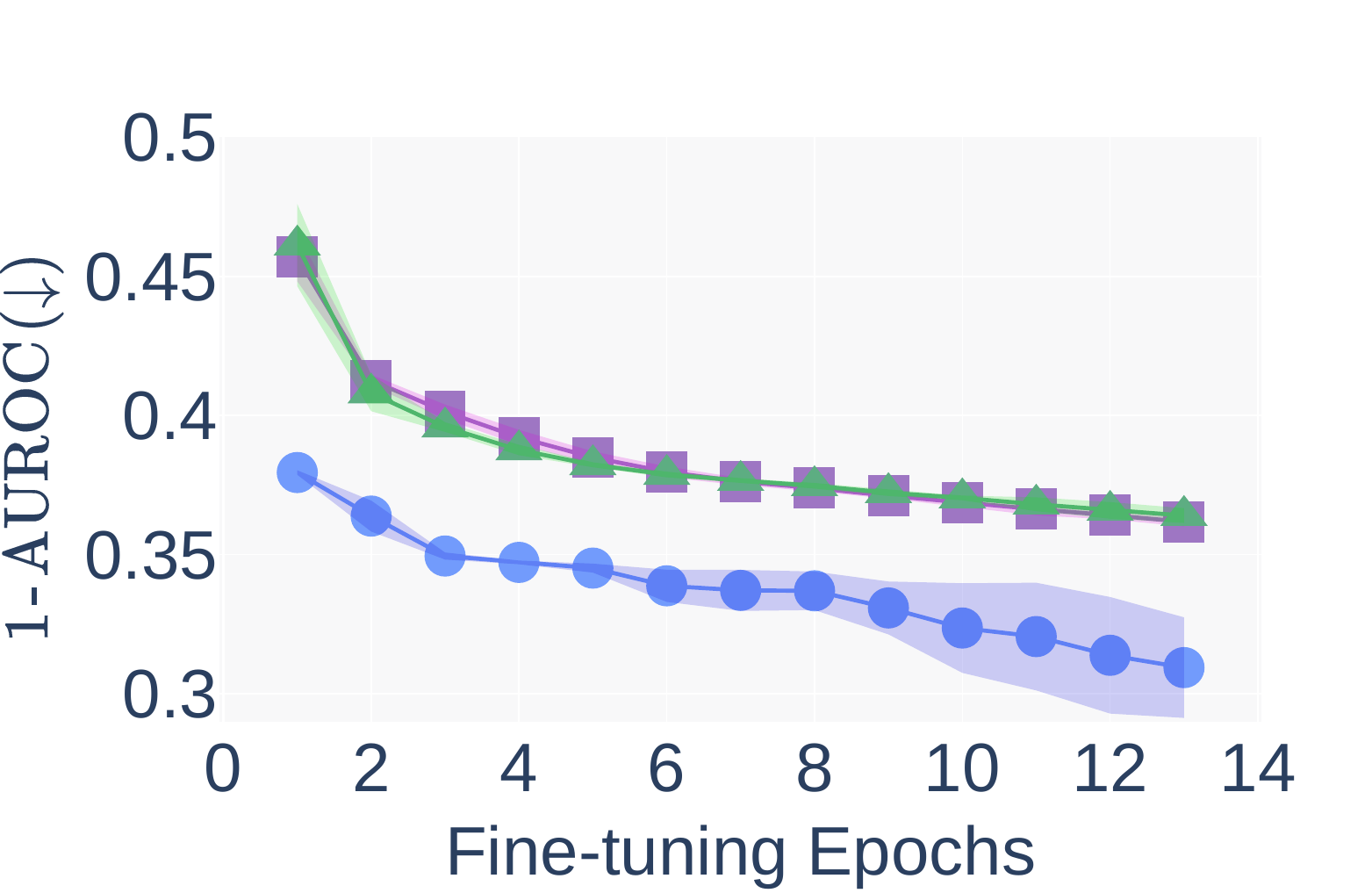} &
        \includegraphics[valign=m,width=0.23\linewidth]{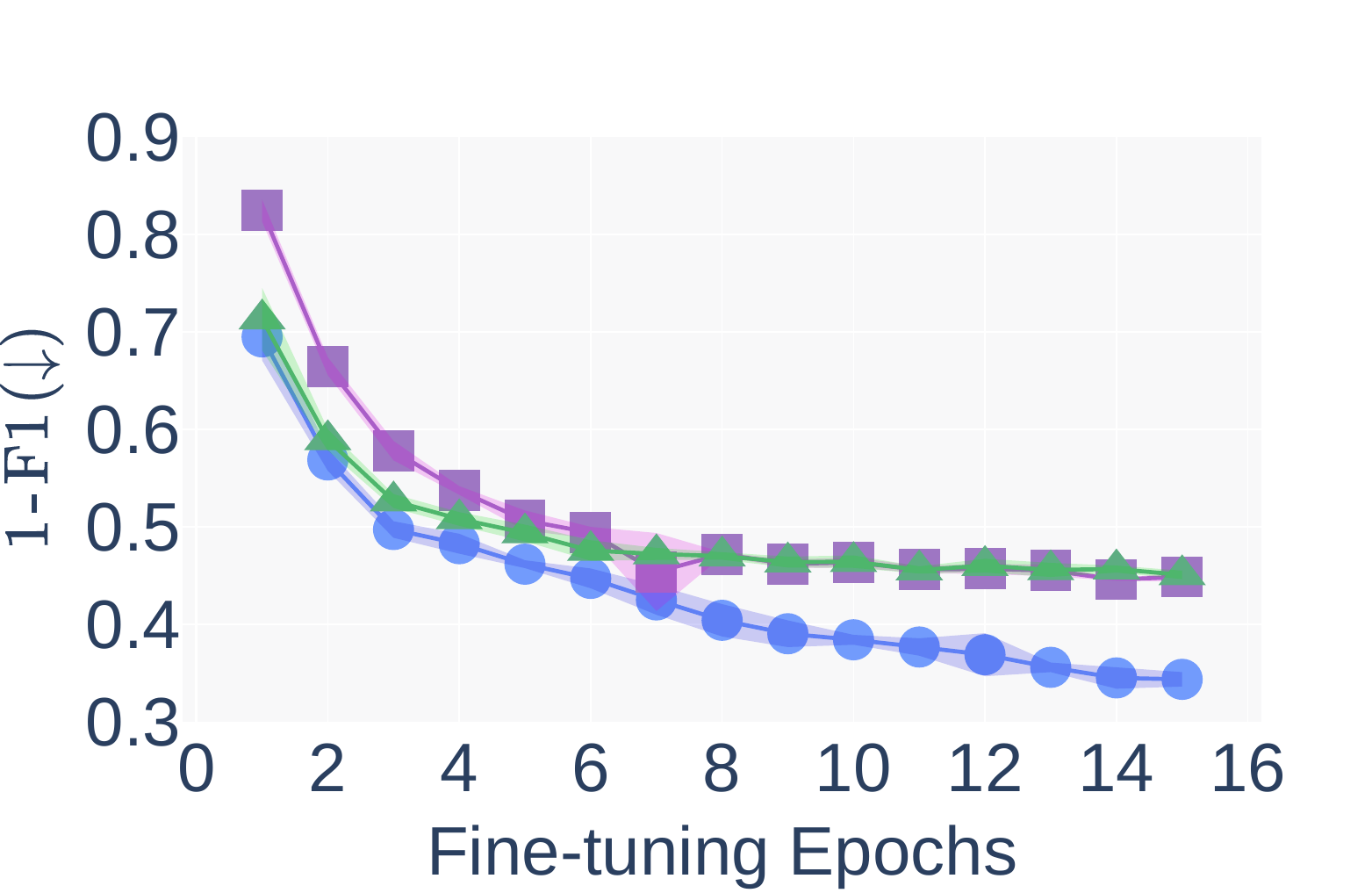}
        \\
        \small{Freezing the embedder} & \includegraphics[valign=m,width=0.23\linewidth]{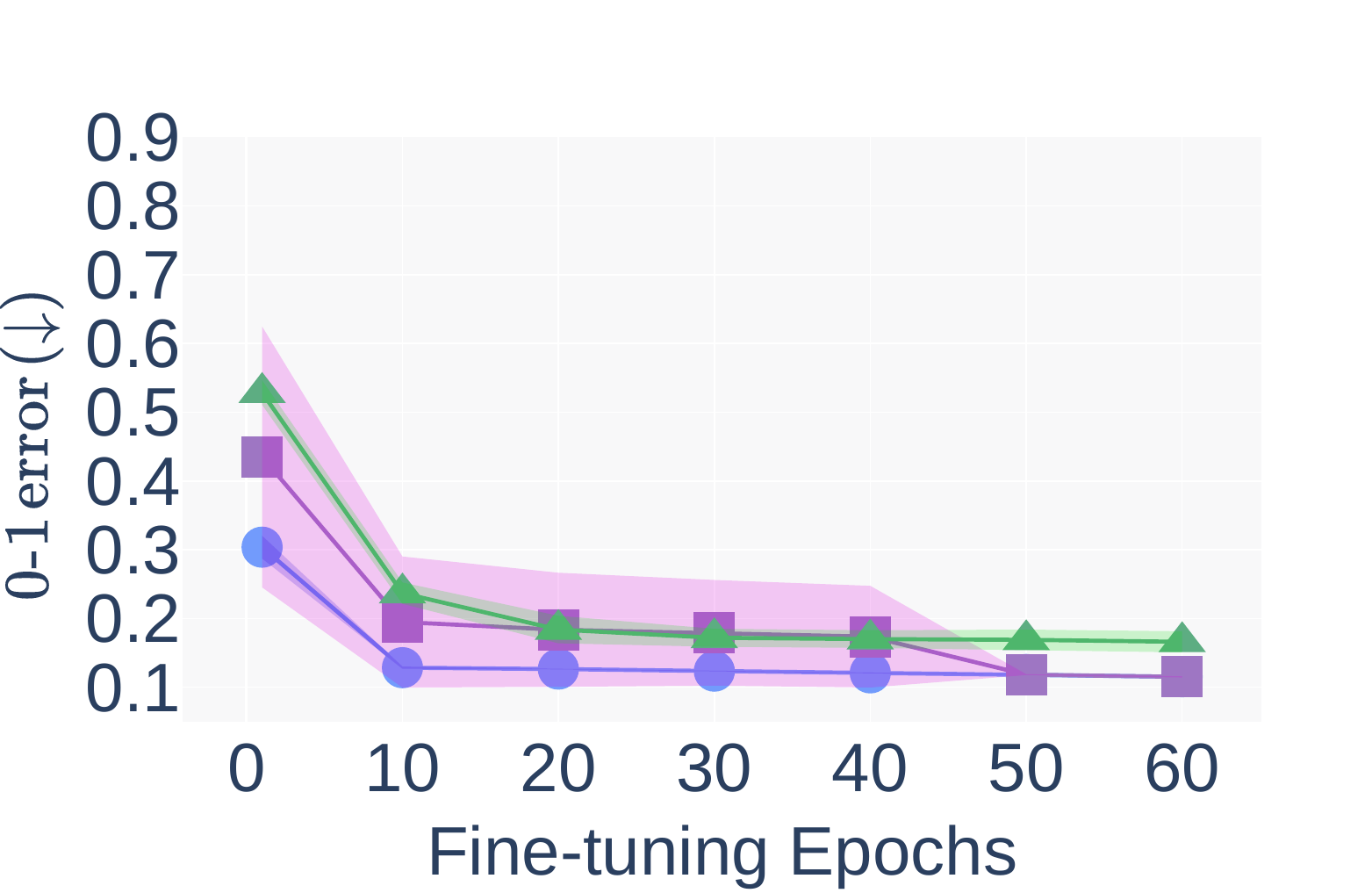} &
        \includegraphics[valign=m,width=0.23\linewidth]{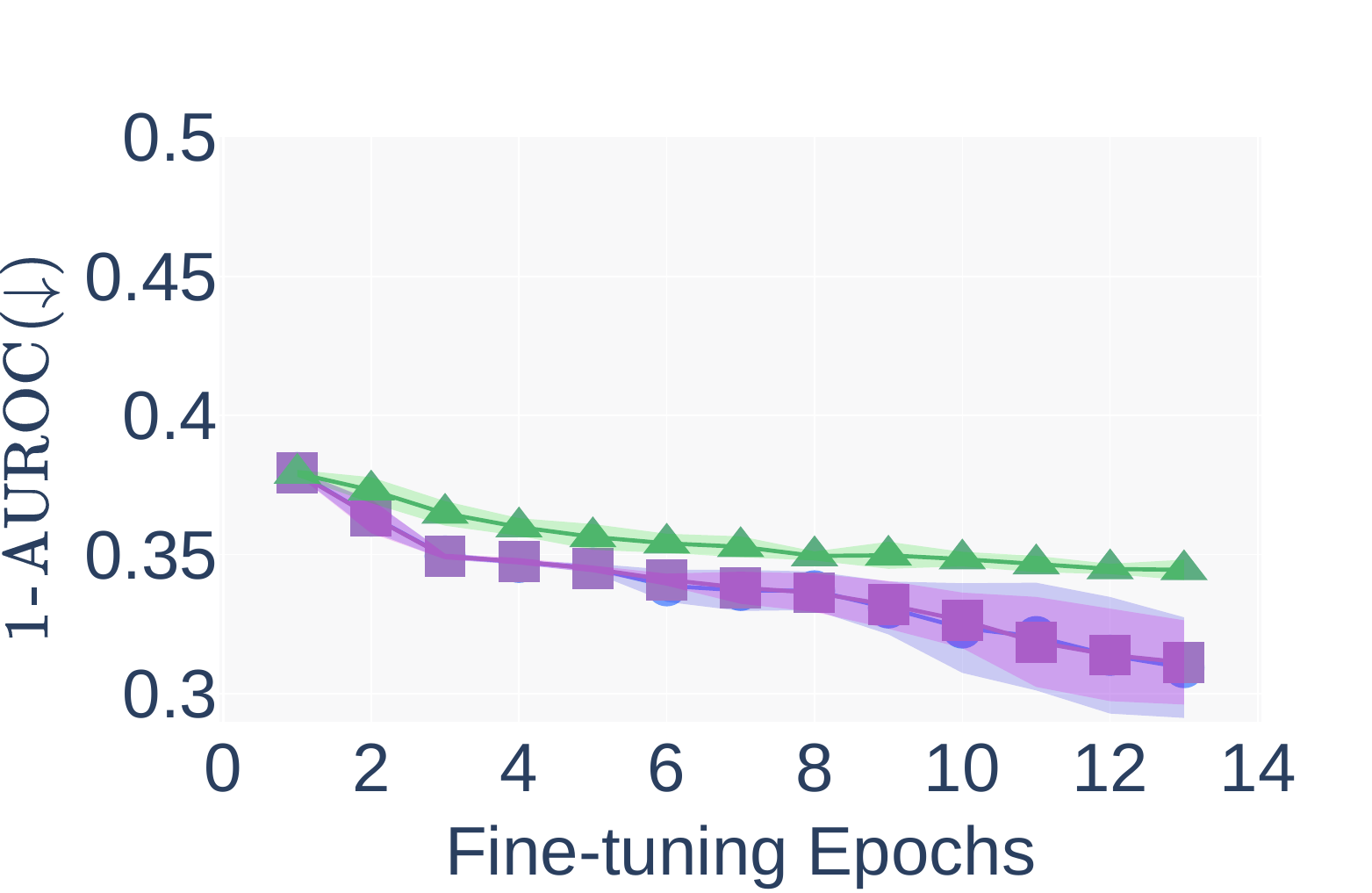} & 
        \includegraphics[valign=m,width=0.23\linewidth]{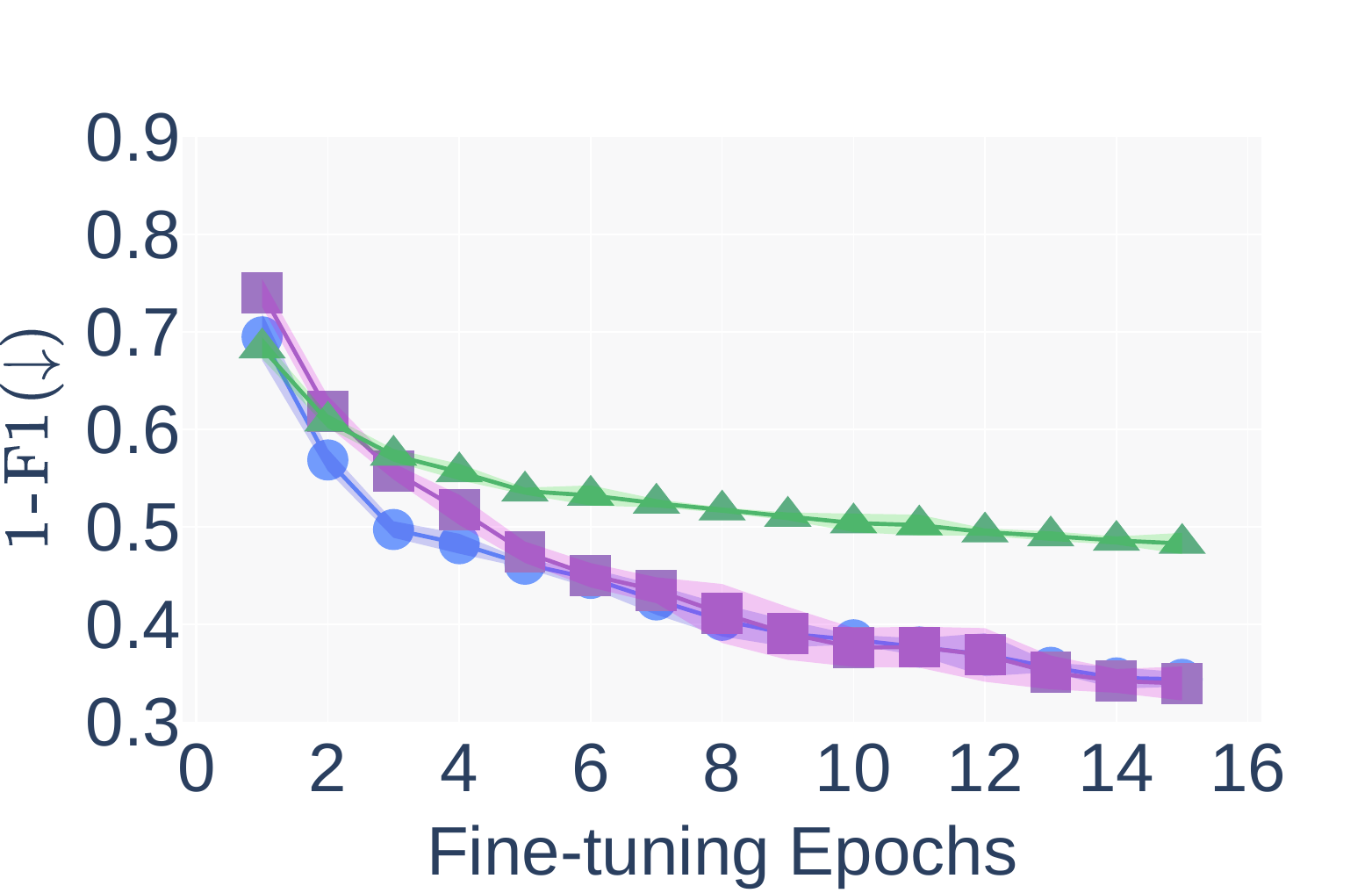}\\
        \end{tabular}
  \centering\caption{Freezing just the embedder, just the model, or both, before full fine-tuning. We also evaluate the impact of training vs. not training the embedder before freezing.}
    \label{fig:frozen}
\end{figure*}

\begin{figure*}[th!]
     \centering
     \begin{subfigure}[b]{0.3\linewidth}
         \centering
         \includegraphics[width=\linewidth]{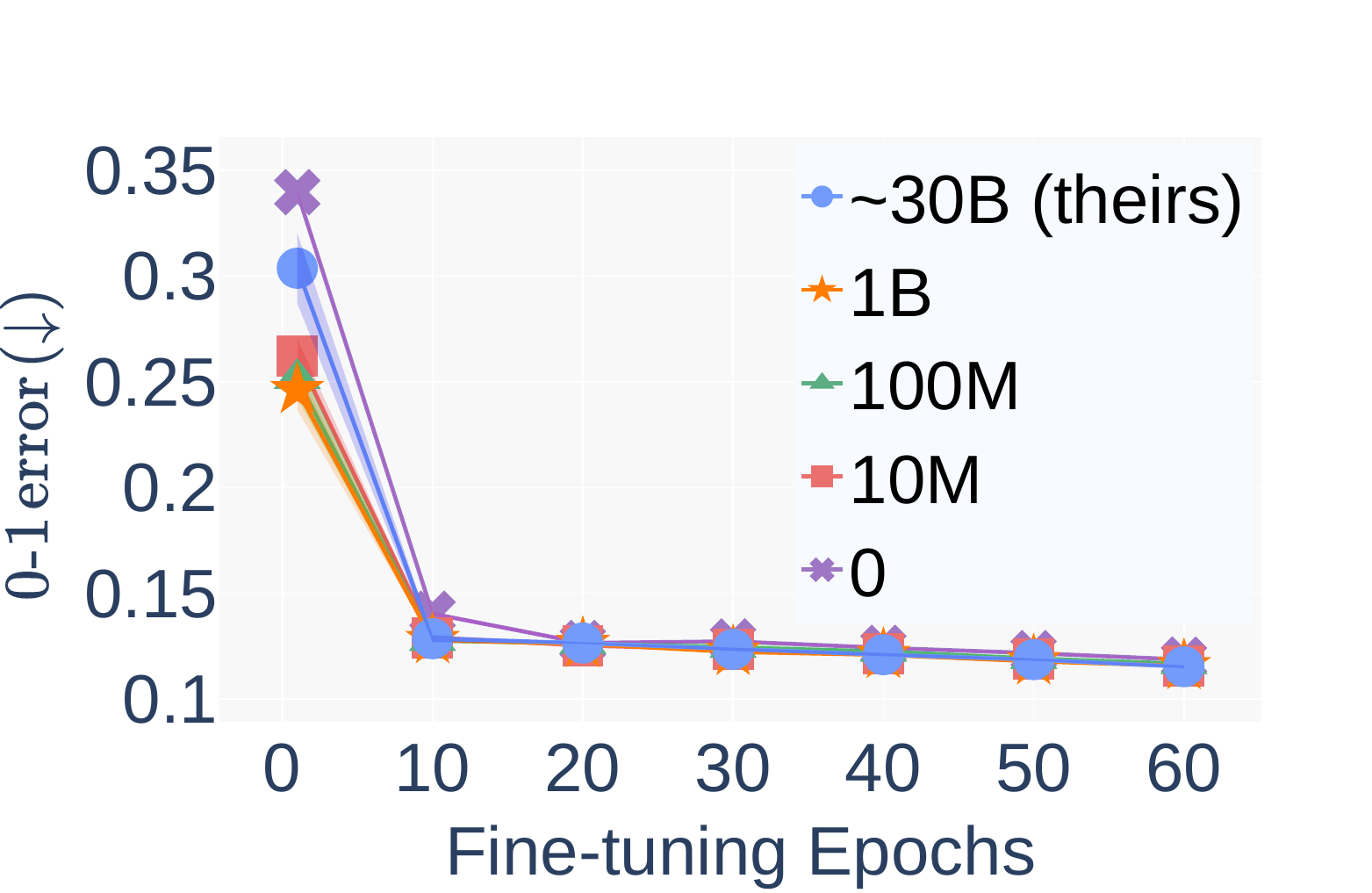}
         \caption{Satellite}
         \label{fig:satellitemini}
     \end{subfigure}
     \hfill
     \begin{subfigure}[b]{0.3\linewidth}
         \centering
         \includegraphics[width=\linewidth]{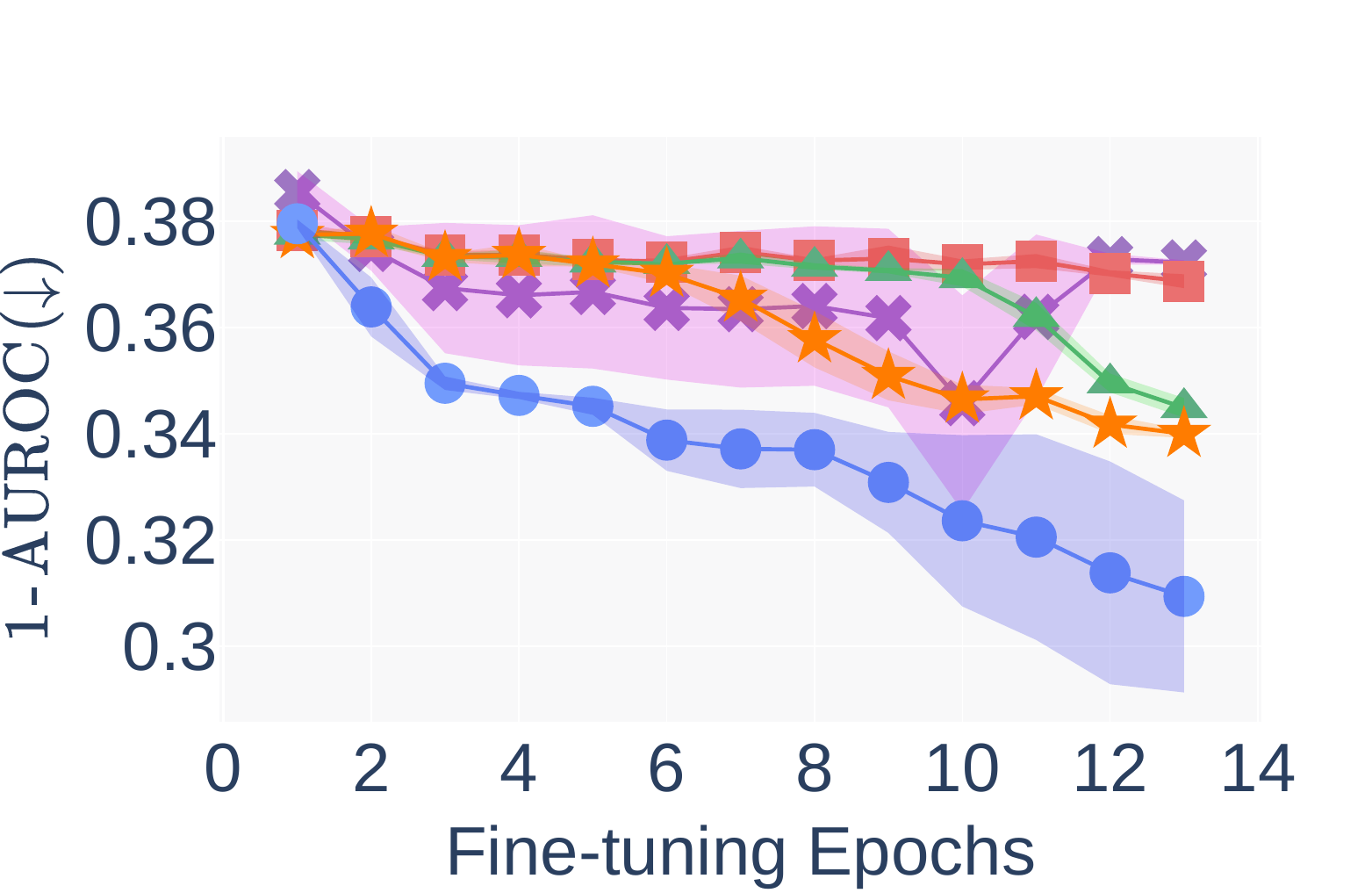}
         \caption{DeepSEA}
         \label{fig:deepseamini}
     \end{subfigure}
     \hfill
     \begin{subfigure}[b]{0.3\linewidth}
         \centering
         \includegraphics[width=\linewidth]{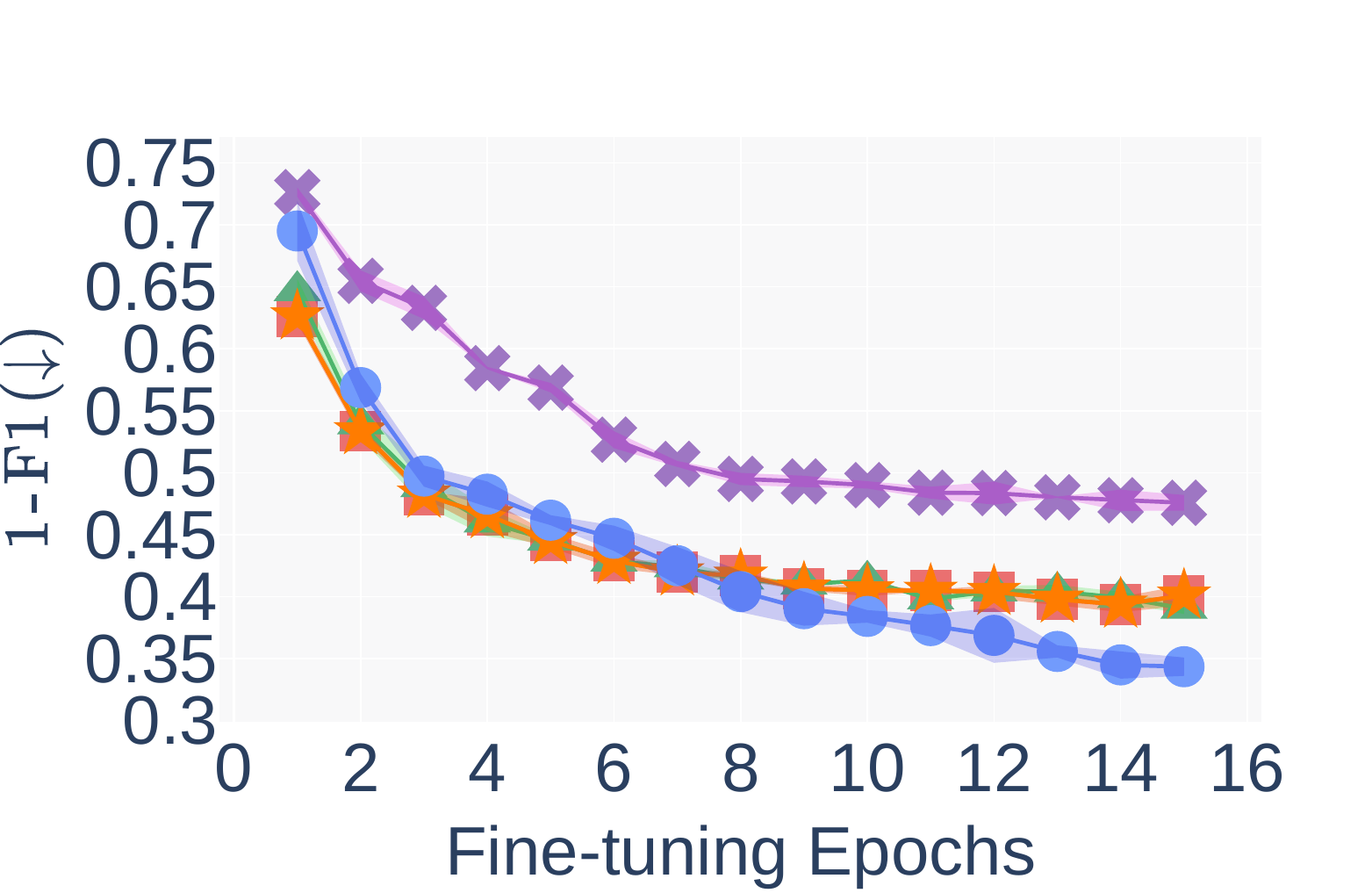}
         \caption{ECG}
         \label{fig:ecgmini}
     \end{subfigure}
        \caption{Effect of different amounts of pre-training data on downstream performance.}
        \label{fig:minirobertas}
\end{figure*}

To better understand how the fine-tuning phase affects the multiple components of ORCA, we experiment with freezing different parts of the pipeline: the embedder, the pre-trained model, or both. We compare our results with the original setup.

Row 1 of \Cref{fig:frozen} shows the results of freezing both the embedder and the pre-trained model, and only fine-tuning the predictor. Across all datasets, the frozen versions perform much worse than the original setup, regardless of embedder training.
This indicates that these datasets are not simple enough to be solved by training a simple predictor.

In row 2, we freeze only the pre-trained model, but fine-tune the embedder and the predictor. These frozen versions also perform much worse than the original setup, indicating that \textbf{fine-tuning the pre-trained model is a critical component of ORCA}, regardless of dataset and embedder training.

Finally, in row 3, we only freeze the embedder, allowing the fine-tuning stage to affect both the model and the predictor. As we already saw in Figure \ref{fig:proxydatasets}, training the embedder is important across all three datasets. However, once this training is done, even if it is frozen, adapting the pre-trained model is sufficient for good task performance. This shows that \textbf{while training the embedder is important for ORCA's success on these datasets, it need not be fine-tuned beyond that}.

\section{Pre-training is not always necessary}
\label{sec:how-much-pretrained-knowledge}

Our previous results show that fine-tuning the model is necessary for good downstream task performance, but they do not show whether using \textit{pre-trained} models is necessary for this.
To answer this question, we use RoBERTa models pre-trained on different amounts of English data.
Specifically, we compare the original RoBERTa-base model to a randomly initialized model with no training data, along with three variants trained on less data~\citep{warstadt-etal-2020-learning}, as shown in \Cref{sec:roberta-details}.

\Cref{fig:minirobertas} shows that performance varies widely depending on the dataset. For Satellite, all models perform the same, showing that the task is simple enough to be solved even without pre-training.
With DeepSEA and ECG, on the other hand, pre-training data on the scale of 30B tokens results in clearly better performance.
These results highlight the importance of comparing to a no pre-training baseline, for ORCA---and indeed all cross-modal fine-tuning work---to ensure that pre-training is actually necessary for the success of the method.

Until the 30B data scale, however, DeepSEA performance remains within the variance of simply fine-tuning a randomly-initialized model, whereas ECG does benefit from even a small amount of pre-training. This shows that even for non-trivial tasks, \textbf{the amount of pre-training has a noticeable effect only at certain scales}.

\section{Conclusion}

We perform a series of ablations to investigate how the different components of ORCA, a recently-proposed method for cross-modal fine-tuning, affect its performance.
Contrary to the original results, we find that embedder training does not help 2D tasks at all, compared to just fine-tuning without training the embeddder.
In 1D tasks, some amount of embedder training is necessary, but unlike the claim in the original paper, more embedder training can even hurt performance on the target task.
In a series of experiments where we freeze components of the ORCA pipeline, we find that fine-tuning the model is crucial for good task performance.
It is not necessary, however, to further train the embedder after stage two.
Finally, we find that for one of the 1D tasks, using a pre-trained model is actually not necessary, indicating the importance of no pre-training baselines in evaluations of cross-modal transfer.

\bibliography{custom}

\appendix

\clearpage
\newpage

\section{Limitations}

\paragraph{Choice of datasets.} We only experiment with three 2D datasets and three 1D datasets, and we do not consider the experiments from the original paper on tabular data, where our findings may not hold. Additionally, due to the widely varying patterns we find in our results, we believe that this is not sufficient for our findings to generalize beyond these specific datasets to the modalities that they represent. This points to a limitation of cross-modal fine-tuning work in general, which would benefit from a larger set of datasets, and in particular, more challenging tasks, as we find that the Satellite dataset is very simple.

\paragraph{Choice of pre-trained models.} Our experiments focus on 1D tasks, for which we only experiment with encoder-only architectures (specifically RoBERTa-type models) even though other encoder-only models and even other architectures (e.g., encoder-decoder and decoder-only models) could also be used. We caution against claims about generalization of our results for these tasks to pre-trained models beyond just RoBERTa.

\paragraph{Ablating stage one.} Our experiments focus on stages two and three of the ORCA pipeline, but stage one, i.e., the creation of the task-specific embedder and predictor, is not something we vary. In \citet{shen2023cross} and in our work, the task-specific embedder consists of a convolutional layer, a layer norm, and a positional embedding, and the predictor consists of a linear projection. It would be interesting to test a much simpler method of converting dimensions in the embedder than a convolutional architecture, e.g., a linear projection, which we leave to future work.

\paragraph{Evaluating what is being transferred.} In Section \ref{sec:frozen}, we show that pre-training is necessary for some cross-modal transfer, but we still do not know exactly what is being transferred. The cross-modal transfer literature posits that pre-trained knowledge is somehow exploited in downstream tasks, but since we do not know how to quantify ``knowledge'' in this setting, we cannot make this claim. It is just as plausible that models pre-trained on tokens beyond a certain scale find better, more general solutions that are a good initialization for adapting to a new task. One way to further probe the transfer hypothesis would be by limiting the number of parameters that are allowed to change during fine-tuning, e.g., by using parameter-efficient fine-tuning with LoRA. We leave an exploration of this to future work.

\section{Dataset details}
\label{sec:dataset}

\begin{table*}[h!]
    \centering
    \begin{tabular}{lllllll}
    \toprule
        \bf Dim. & \bf Target dataset & \bf Type & \bf Metric & \bf \# classes & \bf Proxy dataset & \bf \# classes \\
        \midrule
          & NinaPro & Point & 0-1 error ($\downarrow$) & 18 &  &  \\
         2D & CIFAR-100 & Point & 0-1 error ($\downarrow$) & 100 & CIFAR-10 & 10 \\
          & Darcy Flow & Dense & relative $l_2$ ($\downarrow$) & 10 &  &  \\
         \midrule
          & Satellite & Point & 0-1 error ($\downarrow$) & 24 &  & \\
          1D & DeepSEA & Point (multi-label) & 1 - AUROC ($\downarrow$) & 36 & CoNLL-2003\footnotemark & 7 \\
         & ECG & Point & 1 - F1 ($\downarrow$) & 4 & & \\
         \bottomrule
    \end{tabular}
    \caption{Target datasets of each type along with the proxy datasets used for them in ORCA~\citep{shen2023cross}}
    \label{tab:target-datasets}
\end{table*}

\Cref{tab:target-datasets} shows the target and original proxy datasets considered, along with their dimension, type, number of classes, and the metric used to measure target task performance. The tasks are classified into two types, taking into account whether the task's output is a singular prediction (point) or
multiple predictions (dense). The target datasets are described in more detail below. \\

\begin{figure}[htbp]
    \centering
    \includegraphics[width=1\linewidth]{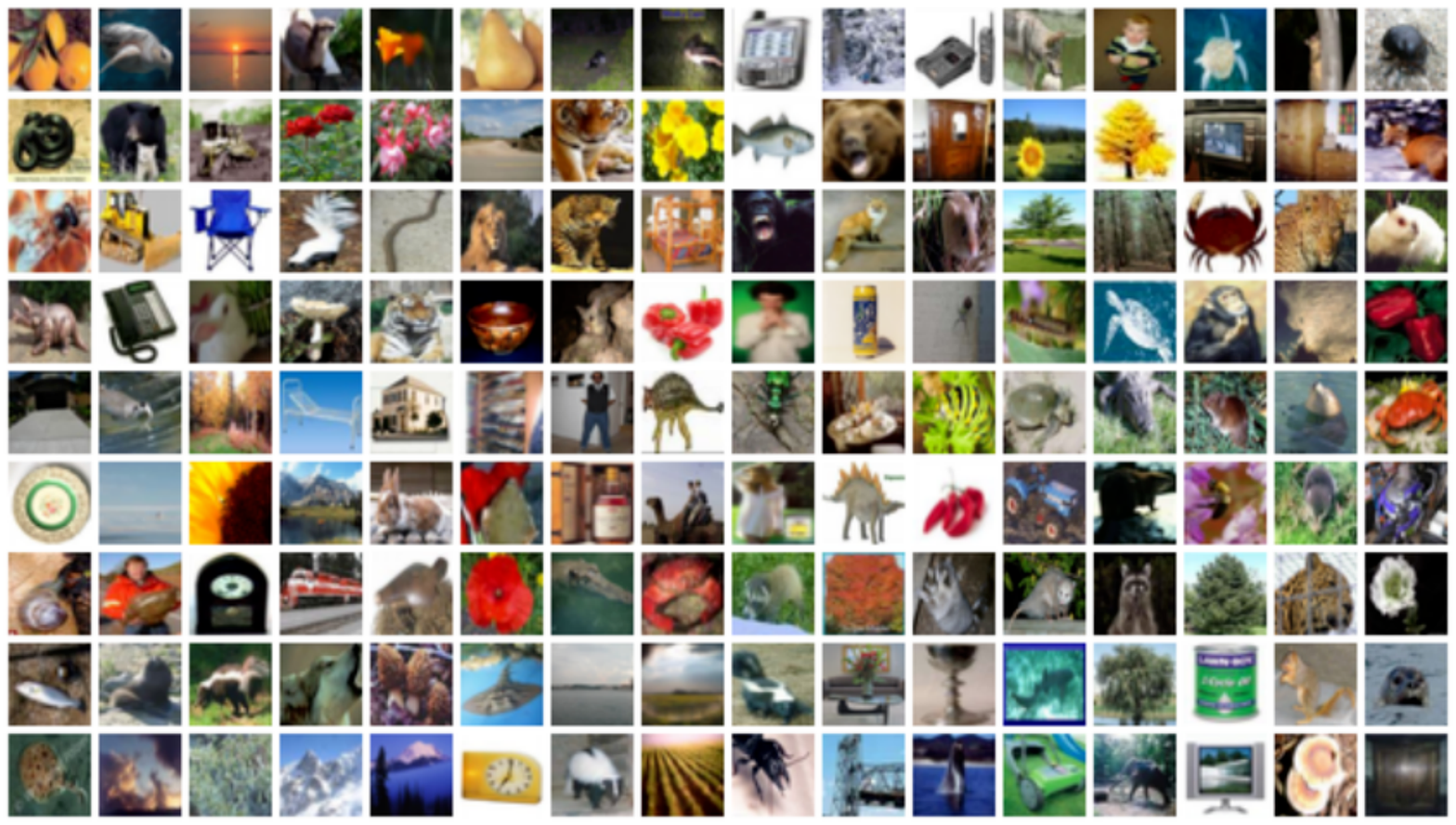}
    \caption{CIFAR-100 examples.}
    \label{fig:cifar100-example}
\end{figure}

\paragraph{CIFAR-100: Standard Image Classification.}

The dataset consists of 32x32 color images divided into 100 classes, based on the object represented by the image. Some examples can be seen in \Cref{fig:cifar100-example}.
\\

\begin{figure}[h]
    \centering
    \includegraphics[width=1\linewidth]{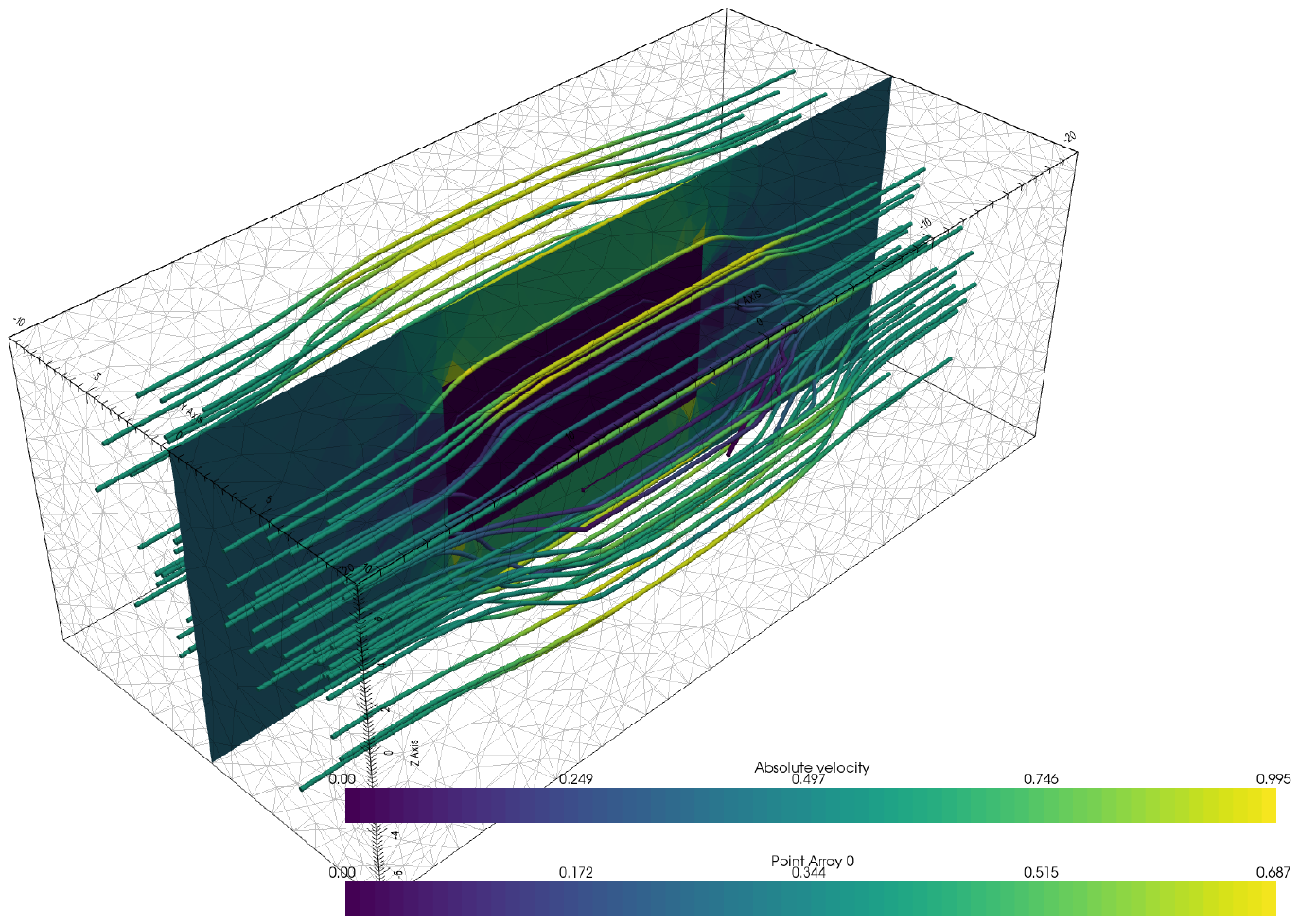}
    \caption{Example from the Darcy Flow dataset.}
    \label{fig:darcy-example}
\end{figure}

\paragraph{Darcy Flow: Solving Partial Differential Equations (PDEs).}

The only regression task considered. Although, for the training stages, the dataset is divided into a total of 10 inferred classes. The dataset consists of 2D grids specifying the initial conditions of a fluid, as an output the same 2D grid on a later time is predicted.
\\

\paragraph{DeepSEA: Predicting Functional Effects From Genetic Sequences.}

The dataset consists of a collection of genomic profiles to estimate the behavior of chromatin proteins, classifying it into 36 classes.
\\

\paragraph{ECG: Detecting Heart Disease.}

\begin{figure}[h]
    \centering
    \includegraphics[width=1.1\columnwidth]{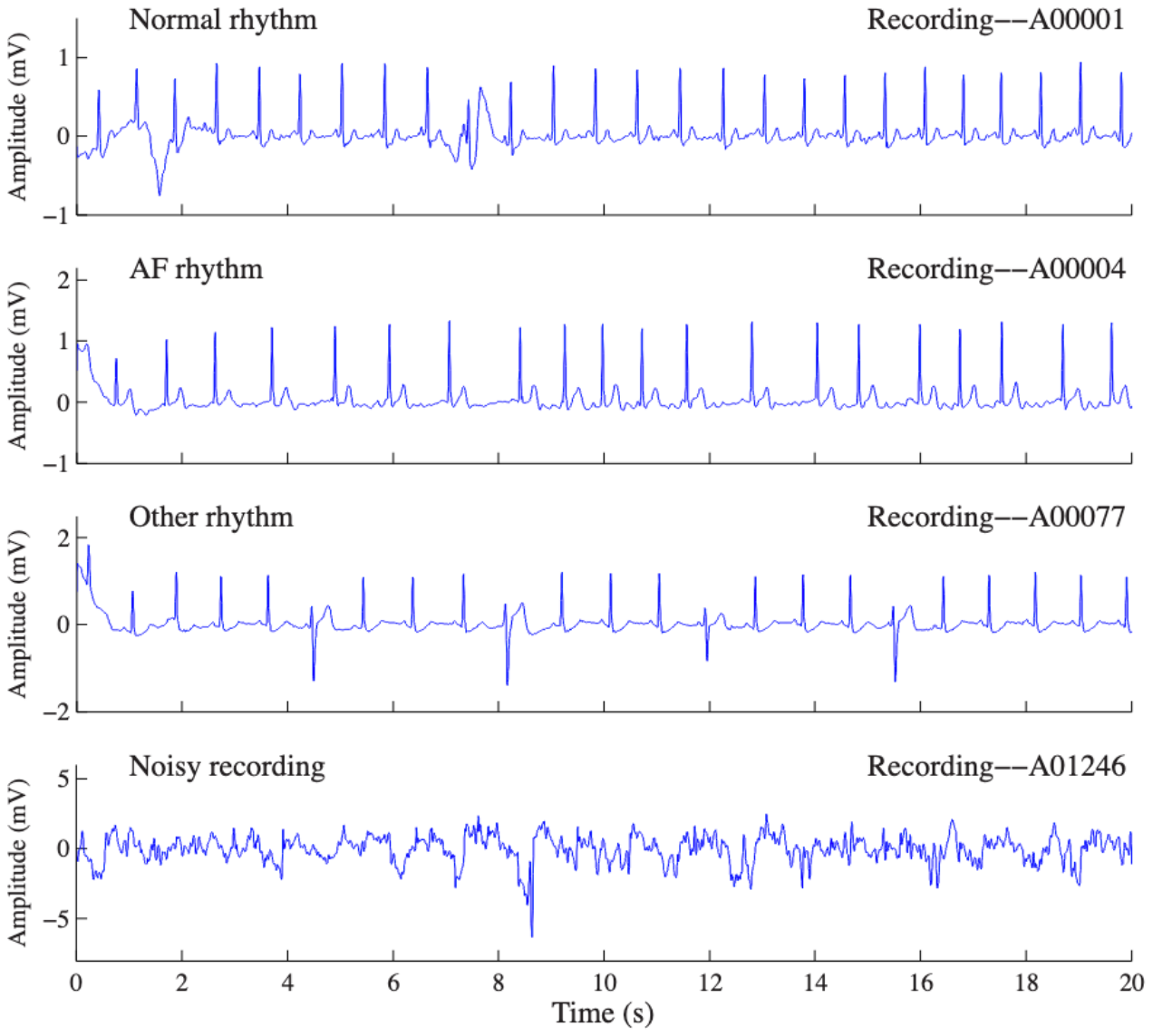}
    \caption{Examples of ECG recordings of the 4 different classes}
    \label{fig:ecg-example}
\end{figure}

The dataset is formed by recordings of up to a minute of Electrocardiograms classified into four classes: normal, disease, other, or noisy rhythms. \Cref{fig:ecg-example} shows an example of each of the classes. 
\\

\footnotetext[3]{We were unable to replicate the exact workflow to create the language features passed to the model, so we used the ones provided in the original ORCA GitHub.}

\begin{figure}[h]
    \centering
    \includegraphics[width=1\linewidth]{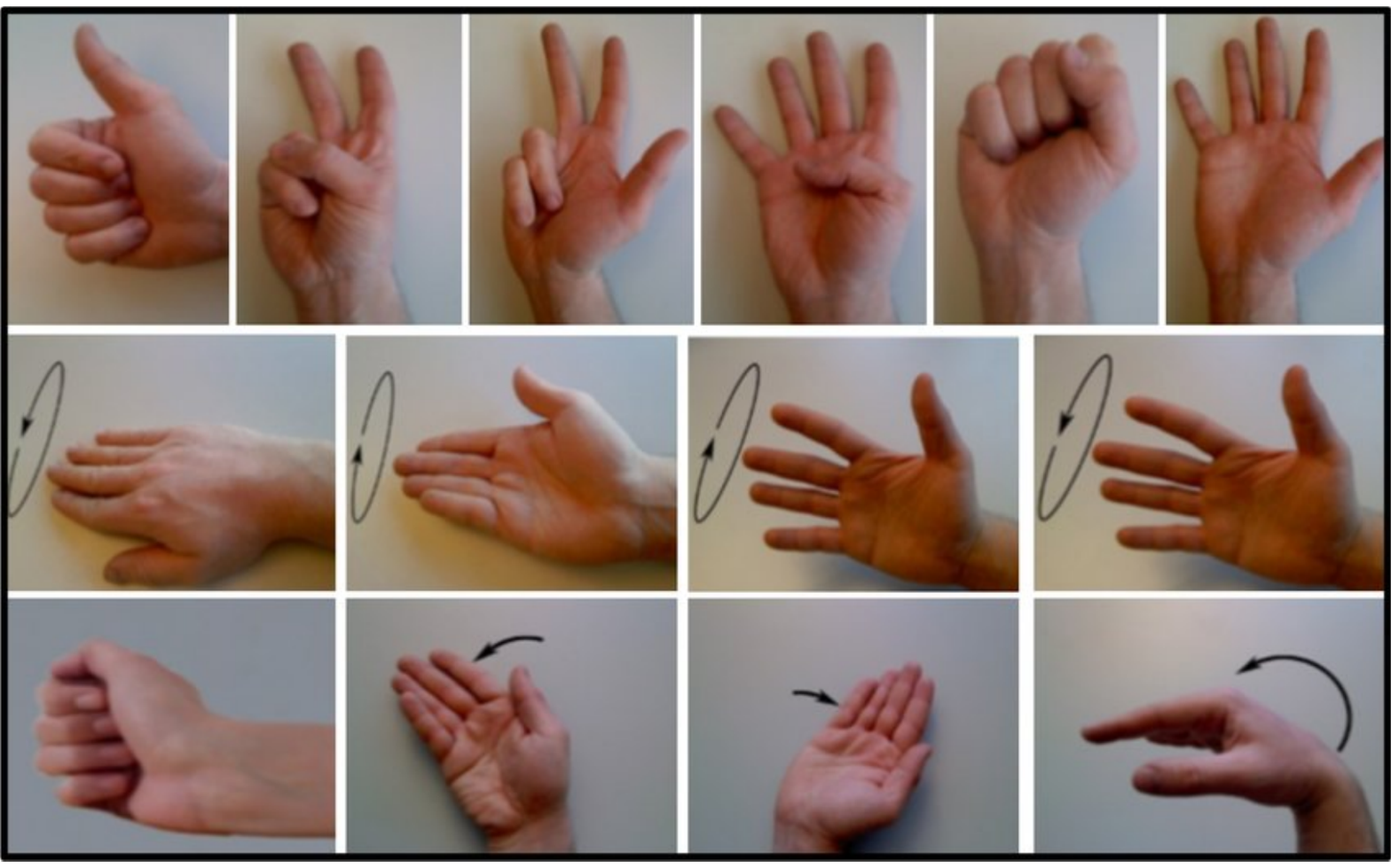}
    \caption{Samples of movements in NinaPro BD5 \cite{shen2019movements}, the dataset contains the electromyography signals of the movements.}
    \label{fig:ninapro-example}
\end{figure}

\paragraph{NinaPro: Classifying Electromyography Signals.}

It takes a subset of NinaPro BD5, to classify the electromyography (sEMG) signals of a collection of hand movements in 18 classes. Some examples of the movements can be seen in Figure \ref{fig:ninapro-example}.
\\

\begin{figure}[h]
    \centering
    \includegraphics[width=1\linewidth]{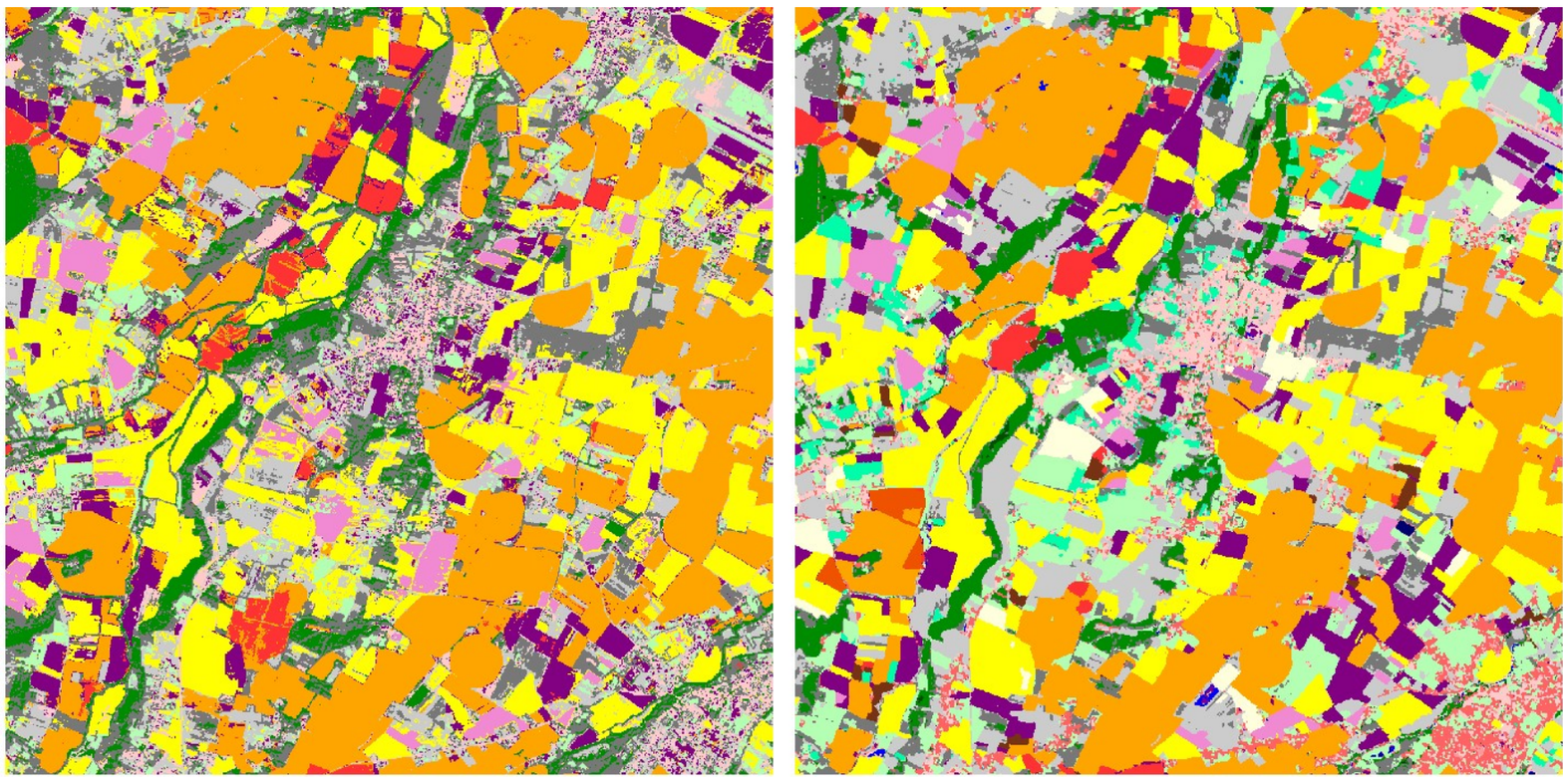}
    \caption{Example of Satellite \cite{petitjean2012satellite}}
    \label{fig:satellite-example}
\end{figure}

\paragraph{Satellite: Satellite Image Time Series Analysis.} 

\begin{algorithm*}[t]
\caption{Efficient approximation of OTDD using class-wise subsampling from \citep{shen2023cross}}\label{alg:otdd}
\textbf{Input:} target dataset $\{x^t, y^t\}$, number of target classes $K^t$, source dataset $S = \{x^s, y^s\}$, subsample size $b$, subsample round $R$
    \begin{algorithmic}
    \For{each class $i \in [K^t]$ in the target dataset} 
        \State Compute class weight $w_i = \frac{number\:of\:target\:data\:in\:class\:i}{total\:number\:of\:target\:data}$
        \State Generate data loader $D_i$ consisting of data in class $i$
    \EndFor
    \For{$i \in [K^t]$}
        \For{$r \in [R]$}
            \State Subsample $b$ target data points $D_{ir}$ uniformly at random from $D_i$
            \State Compute class-wise distance $d_{ir} = OTDD(D_{ir},S)$
        \EndFor
        \State Approximate class-wise OTDD by $d_i = \frac{1}{R} \sum^{R}_{i=1}{d_{ir}}$
    \EndFor
    \State Approximate OTDD by $d = \sum^{K^t}_{i=1}{w_i \dot d_i}$
    \end{algorithmic}
\end{algorithm*}

The dataset consists of satellite image time series (SITS), tracking the land changes over the years, classifying them into 24 land cover types.

\section{Embedder and predictor details}
\label{sec:appendix:embedder-details}

As described in \Cref{fig:ORCA}, in the first stage of the ORCA workflow \cite{shen2023cross}, a task-specific embedder and predictor are created to support any combination of input-output dimensions. Throughout all our experiments, we kept the same architectures used in the original paper, which we will explain in this section.

\paragraph{Task-specific Embedding Network} 

The architecture is composed of a convolutional layer with an input channel of the target dataset and an output channel of the dimension of the pre-trained model embedding space. The kernel size and stride can be treated as a hyperparameter, but in all our experiments for the 2D tasks both are set to four and, for the 1D tasks, are computed based on the input and target sequence length. After this, a layer norm and a positional embedder are added to obtain the final representation. 

\paragraph{Task-specific Predictor} 
Given the diversity of the tasks considered, two different architectures are implemented depending on the target task type. For the point tasks, average pooling along the sequence length dimension is applied, to obtain 1D tensors with the same length as the dimension of the pre-trained model embedding space. Then to map to the number of classes of the target dataset, a linear layer is used. For dense tasks, a linear layer is applied to the sequence outputs to adjust the tensor shape. Then, this tensor is molded to the desired output dimension.

\section{OTDD approximation implementation}

Following the original ORCA implementation \cite{shen2023cross}, we also used an approximation of OTDD using class-wise subsampling, as described in \Cref{alg:otdd}. 

As described in the original paper, to tackle potential memory issues when computing OTDD, the dimensionality of the feature vectors is reduced by taking the average along the sequence length dimension. On top of that, the target dataset is divided into subsets based on the labels, each of these subsets will be approximated with the average of batch samples (the number of maximum samples taken from each class is determined for every dataset). Then the OTDD between each class representative and a sample of the proxy dataset (5000 samples for CIFAR-10 and 2000 for CONLL 2003) is computed. Finally, the overall OTDD is approximated by the weighted sum of the OTDD of all the classes in the task dataset. 

\section{Details on pre-trained RoBERTa models}
\label{sec:roberta-details}

\Cref{tab:minirobertas} provides information about the amount of training data seen by the different RoBERTa variants released by \citet{warstadt-etal-2020-learning}.

\begin{table}[h]
    \centering
    \begin{tabular}{lll}
     \toprule
        \textbf{Model} & \textbf{Training data} \\
         \midrule
        roberta-base & \textasciitilde{}30B \\
        roberta-base-1B-2 & 1B \\
        roberta-base-100M-3 & 100M \\
        roberta-base-10M-3 & 10M \\
        roberta-base-random & 0 \\
         \bottomrule
    \end{tabular}
    \caption{Models for pre-trained knowledge comparison, and their training data in number of tokens.
    }
    \label{tab:minirobertas}
\end{table}

\end{document}

%% file: macros.tex
\usepackage{todonotes}
\usepackage[normalem]{ulem}

\makeatletter
\newcommand*\iftodonotes{\if@todonotes@disabled\expandafter\@secondoftwo\else\expandafter\@firstoftwo\fi}  
\makeatother

